\documentclass[10pt,journal,compsoc]{IEEEtran}
\ifCLASSOPTIONcompsoc
  \usepackage[nocompress]{cite}
\else
  \usepackage{cite}
\fi

\usepackage{graphicx}
\usepackage{subfigure}
\usepackage{multirow}
\usepackage{amssymb}
\usepackage{amsmath}
\usepackage{booktabs}
\usepackage[ruled]{algorithm2e}
\usepackage{color}
\usepackage{ragged2e}
\usepackage{hyperref}
\usepackage{cite}

\usepackage{diagbox}

\ifCLASSINFOpdf
\else
\fi
\usepackage{algorithmic}
\begin{document}
%
\title{Prioritized Subnet Sampling for Resource-Adaptive Supernet Training}
%
%
%
%

\author{
        Bohong Chen,
        Mingbao Lin,
        Rongrong Ji,~\IEEEmembership{Senior Member,~IEEE},
        Liujuan Cao

\IEEEcompsocitemizethanks{


\IEEEcompsocthanksitem B. Chen, R. Ji and L. Cao (Corresponding  Author) are with the Media Analytics and Computing Laboratory, School of Informatics, Xiamen University, Xiamen 361005, China (e-mail: caoliujuan@xmu.edu.cn).\protect
\IEEEcompsocthanksitem M. Lin is with the Tencent Youtu Lab, Shanghai 200233, China.\protect
\IEEEcompsocthanksitem R. Ji is also with Institute of Artificial Intelligence, Xiamen University, Xiamen 361005, China.\protect
}

\thanks{Manuscript received April 19, 2005; revised August 26, 2015.}}

%
%

\markboth{IEEE TRANSACTIONS ON PATTERN ANALYSIS AND MACHINE INTELLIGENCE Under Review}
{Shell \MakeLowercase{\textit{et al.}}: Bare Demo of IEEEtran.cls for Computer Society Journals}
%



\IEEEtitleabstractindextext{%
\begin{abstract}
\justifying
A resource-adaptive supernet adjusts its subnets for inference to fit the dynamically available resources. In this paper, we propose prioritized subnet sampling to train a resource-adaptive supernet, termed PSS-Net. We maintain multiple subnet pools, each of which stores the information of substantial subnets with similar resource consumption. Considering a resource constraint, subnets conditioned on this resource constraint are sampled from a pre-defined subnet structure space and high-quality ones will be inserted into the corresponding subnet pool. Then, the sampling will gradually be prone to sampling subnets from the subnet pools. Moreover, the one with a better performance metric is assigned with higher priority to train our PSS-Net, if sampling is from a subnet pool. At the end of training, our PSS-Net retains the best subnet in each pool to entitle a fast switch of high-quality subnets for inference when the available resources vary. Experiments on ImageNet using MobileNet-V1/V2 and {\color{black}ResNet-50} show that our PSS-Net can well outperform state-of-the-art resource-adaptive supernets. Our project is publicly available at \url{https://github.com/chenbong/PSS-Net}.

\end{abstract}

}

\maketitle

\IEEEdisplaynontitleabstractindextext

%
\IEEEpeerreviewmaketitle

\IEEEraisesectionheading{\section{Introduction}\label{sec:intro}}

%
%
%
%
\IEEEPARstart
{D}{eep} neural networks (DNNs) have advanced many tasks of artificial intelligence. Traditional DNNs are designed with fixed network structures, where the hardware resources, such as computation ability and memory footprint, have to be allocated in advance to enable the running of DNNs. 
Unfortunately, the equipped resources are greatly different among heterogeneous devices.
Even for the same device, the availability of hardware resources varies over time. These limitations barricade DNNs to be deployed. Though many works obtain a compact network via model pruning~\cite{ruan2021dpfps,wang2020pruning}, network architecture search (NAS)~\cite{cai2019once,dai2019chamnet}, \emph{etc.}, the resulting model structures are still fixed, making it hard to make full use of changeable hardware resources.

By adjusting the runtime subnets to fit the dynamically available resources, designing resource-adaptive supernets has attracted increasing attention from the research community.
According to the subnet sampling strategy during supernet training, traditional resource-adaptive supernets can be generally divided into two groups including uniform subnet sampling and random subnet sampling.

\begin{figure}[!t]
\centering
\includegraphics[width=0.85\columnwidth]{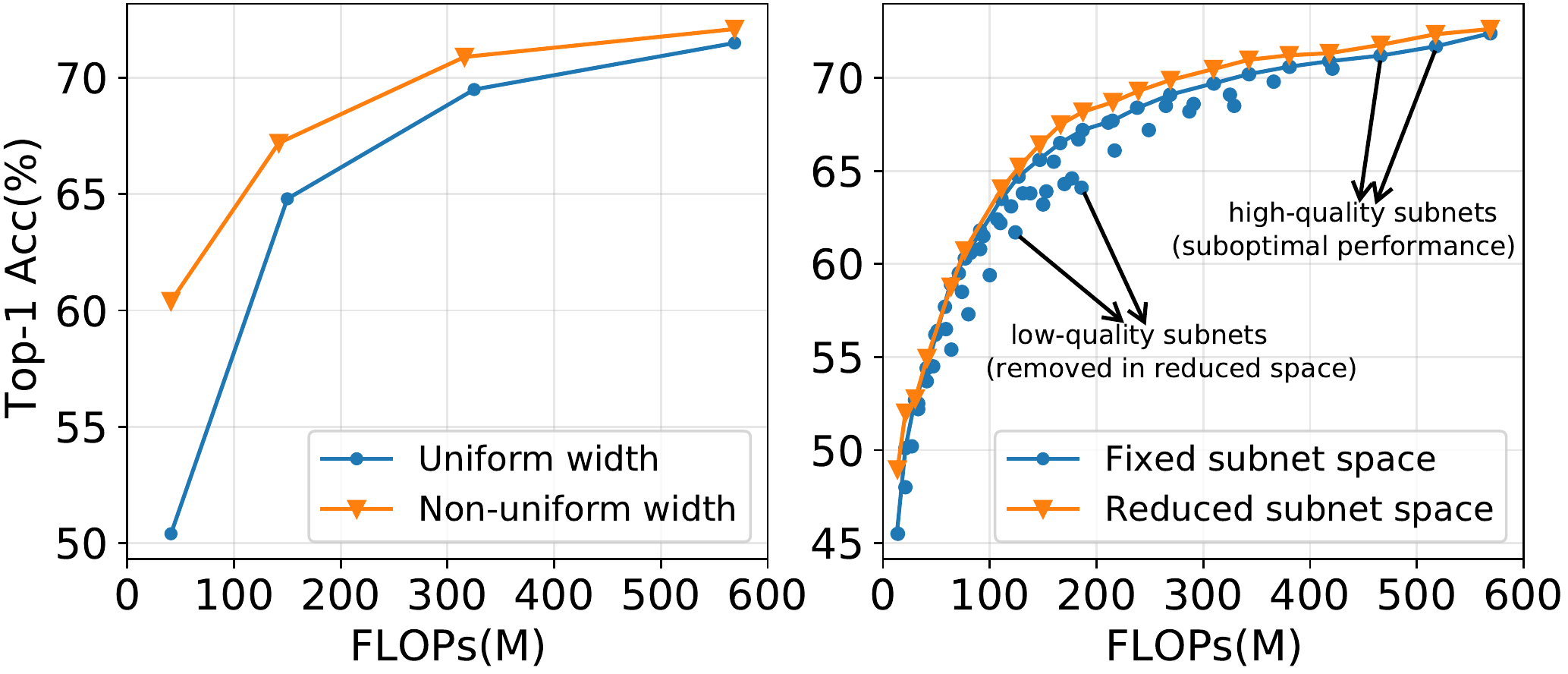}
\vspace{-0.5em}
\caption{\textbf{Left}: Performance of SlimmableNetwork~\cite{yu2018slimmable} with uniform width and non-uniform width~\cite{liu2019metapruning}.
\textbf{Right}: Performance of MutualNet~\cite{yang2020mutualnet} with fixed and reduced search spaces.}
\label{fig:moti}
\vspace{-1.5em}
\end{figure}

{\color{black}
For uniform subnet sampling, some fixed multipliers are applied to all layers to scale the network width, such as USNet~\cite{yu2019universally}, thus providing different subnets at runtime.
Two techniques are introduced to train different subnets in the supernet, including ``sandwich rule'' where three subnets with the maximum, medium, and minimum sizes are sampled to process a batch of images; as well as ``inplace distillation" where the largest subnets are supervised by the ground-truth labels and its outputs guide the learning of medium and smallest subnets.
Besides uniform width, MutualNet~\cite{yang2020mutualnet} also considers the resolution of input images to enhance subnet performance. It also improves the ``inplace distillation" in a style of Kullback-Leibler divergence.
}
As for random subnet sampling, these methods~\cite{cai2019once,lou2021dynamic,yu2020bignas,wang2021attentivenas} randomly sample subnets from a supernet for training in each iteration, and then search for optimal subnets of different sizes from the pre-trained supernet using heuristic algorithms. The learning is also known as one-shot NAS.
However, the performance is far from satisfactory for real-world applications. We attribute it to three open issues, \emph{i.e.}, \textit{homogeneity of subnets}, \textit{search overhead}, and \textit{subnet mismatch}.

In terms of homogeneity of subnets, uniform subnet sampling \cite{yu2018slimmable,yu2019universally,yang2020mutualnet} simply enforces shared multipliers to all layers without investigating the diversity of layer-wise width. Consequently, only one subnet structure can be made given a particular resource constraint such as FLOPs, latency, \emph{etc}. As previously studied in network pruning~\cite{yu2019autoslim,lin2020channel,liu2019metapruning} and NAS~\cite{guo2020single,cai2019once}, the optimal subnet at a particular size often results from the non-uniform sampling. To verify this, we perform a validation experiment on ImageNet by replacing the uniform width of MobileNet-V1~\cite{yu2018slimmable} with a non-uniform form from~\cite{liu2019metapruning}. Fig.\,\ref{fig:moti}(\textbf{Left}) shows a better trade-off between accuracy and FLOPs from the non-uniform strategy. Thus, the homogeneity of subnets, if not well addressed, results in suboptimal performance.

In terms of search overhead, though random subnet sampling may produce different subnet structures over supernet training \emph{w.r.t.} a particular resource, which alleviates the homogeneity of subnets, additional overheads are required to search for the optimal subnet of a particular size. It takes around 40 GPU hours to find out each target subnet in the trained supernet~\cite{cai2019once}. Though this additional cost is acceptable in offline environment, it greatly prevents resource-adaptive supernet from real-time running environment where the available hardware resources frequently vary a lot due to the concurrently running applications.

Subnet mismatch refers to that the sampled subnets during training deviate considerably from the target subnets in inference~\cite{yang2020mutualnet,cai2019once,lou2021dynamic}. Yang \emph{et al}.~\cite{yang2020mutualnet} observed that only around 30\% of sampled subnets fall into the trade-off curve of accuracy \emph{vs.} FLOPs. This indicates that amounts of computation is wasted on training these low-quality subnets that would not be considered in inference. Besides, without sufficient training, these high-quality subnets may result in suboptimal performance in the real-world applications as shown in Fig.\,\ref{fig:moti}(\textbf{Right}). Thus, it is necessary to gradually reduce the subnet structure space during the network training, so as to concentrate more on these high-quality subnets~\cite{sahni2021compofa}.

\begin{figure}[!t]
\centering
\includegraphics[width=0.75\columnwidth]{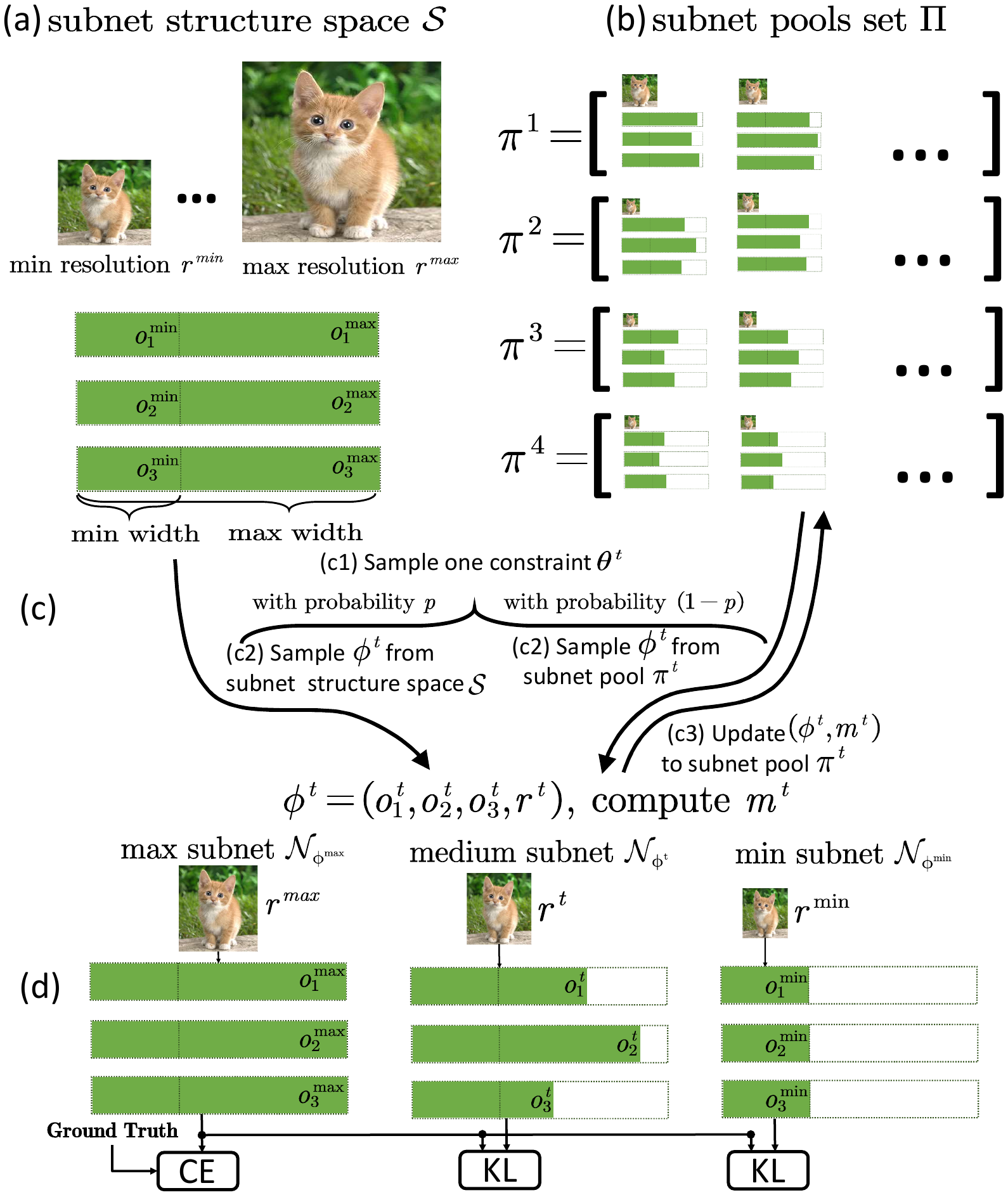}
\vspace{-0.5em}
\caption{PSS-Net training framework. 
(a) Initializing the subnet structure space $\mathcal{S}$. (b) Setting up a set of subnet pools $\Theta = \{\pi^1, \pi^2, ..., \pi^T\}$.
(c) Given a resource constraint $\theta^t$, sampling a medium-sized subnet structure $\phi^t$ either from the subnet structure space $\mathcal{S}$, or from the corresponding subnet pool $\pi^t$.
(d) Slimmable training for $\mathcal{N}_{\phi^{max}}, \mathcal{N}_{\phi^{t}}, \mathcal{N}_{\phi^{min}}$.
The details are described in Sec.\,\,\ref{sec:methodology} and Alg.\,\ref{alg:PSS-Net}.
}
\label{fig:framework}
\vspace{-1.0em}
\end{figure}

Overall, great challenges still remain in training a resource-adaptive supernet. On one hand, we pursue high-performing DNNs for practical deployment. On the other hand, the deployed model should be switchable instantly if the available hardware resources are changed. Reflecting upon the subnet mismatch problem mentioned above, we can know that uniform subnet sampling and random subnet sampling fail to accurately model subnets towards the available resources. A tedious search process is executed to find out the target subnet when the available resources vary.

To solve these problems, we propose prioritized subnet sampling to train a resource-adaptive supernet, termed PSS-Net, as illustrated in Fig.\,\ref{fig:framework}. 
We first build a subnet structure space by applying varying multipliers to different layers, which eliminates the homogeneity of subnets resulting from uniform subnet sampling. 
Then, we set up a group of subnet pools, each of which stores the information of these high-quality subnets conditioned on the same resource constraint. 
In the early training batches of PSS-Net, we sample a medium-sized subnet from a pre-built subnet structure space according to the structure distribution~\cite{wang2021attentivenas}.
To measure the quality of a sampled subnet, we consider its batch loss as the performance metric, alternative to validation accuracy which requires a computationally heavy subnet training. We update the performance metric using moving average if the same subnet structure is sampled multiple times. For each high-quality subnet, its information is inserted into the corresponding subnet pool.
When the subnet pool is full, subnet sampling will gradually shift from the subnet structure space to each subnet pool to ensure that high-quality subnets are entitled with sufficient training, which solves subnet mismatch. Moreover, when sampling is from a subnet pool, the one with a better performance metric than other high-quality subnets is given with higher priority to being sampled to train PSS-Net.
At the end of PSS-Net training, we only need to preserve the best subnet in each pool. Thus, PSS-Net results in a set of subnets that can be alternatively deployed to adapt to the currently available resources, without additional overhead for subnet search.
PSS-Net introduces the subnet pools to collect high-quality subnets during training, which eliminates the search consumption. This implies the idea of trading space for time.

We use the MobileNet-V1~\cite{howard2017mobilenets}/V2~\cite{sandler2018mobilenetv2} {\color{black}and ResNet-50~\cite{he2016deep}} as our supernets, and the experiments are conducted on ImageNet {\color{black}and COCO}. 
{\color{black}Note that the performance increase on MobileNets is challenging for their light-weight designs. Nevertheless, the} results show that our PSS-Net significantly outperforms SOTAs
under different constraints.

{\color{black}
USNet~\cite{yu2019universally} and MutualNet~\cite{yang2020mutualnet} share similarity with our study in supernet training to obtain subnets of different sizes, and switch these subnets to achieve resource-adaptive inference. The difference includes: 
1) USNet and MutualNet apply the same width multiplier to all layers and obtain uniform subnet width. Our PSS-Net adopts layer-wise widths and introduces subnet pooling for high-quality subnet collection. 
2) Our PSS-Net can obtain optimal subnets constrained under different metrics such as FLOPs, Parameters, GPU/CPU latency, \textit{etc.}, which is inapplicable for USNet and MutualNet.
To train PSS-Net, we borrow ``sandwich rule'' and ``inplace distillation'' from previous USNet~\cite{yu2019universally} and MutualNet~\cite{yang2020mutualnet}. 
Our major contributions include:
\begin{itemize}
    \item We propose a subnet pooling-based resource-adaptive supernet training that can sample and train subnets for different resource constraints and adjust them adaptively during the inference time.
    \item We collect subnets by introducing subnet pools during training to obtain high-quality subnet structures with different resource constraints. It eliminates subnet search process after supernet training.
    \item The experimental performance on classification task and object detection task demonstrates the effectiveness of our proposed PSS-Net.
\end{itemize}
}

\section{Related Work}

\textbf{Compact Networks}. To overcome the hurdle of deploying models on resource-limited devices, many previous works focus on obtaining compact networks, typically through model compression \cite{guo2020channel,ruan2021dpfps,wang2020pruning,li2020eagleeye,liu2019metapruning}, network architecture search (NAS) \cite{wang2021attentivenas,cai2019once,he2018amc}, \emph{etc}. Network pruning, a form of model compression, removes redundant weights/filters to satisfy resource constraints in general hardware. Han \emph{et al.}~\cite{han2015learning} proposed to recursively remove small-magnitude weights and retrain the $\ell_2$-regularized subnetwork to recover accuracy. Lin~\emph{et al.}~\cite{lin2020hrank} discarded filters with low-rank feature map outputs. The lottery ticket hypothesis~\cite{frankle2018lottery} randomly initializes a dense network and trains it from scratch. The subnets with high-weight values are extracted, and retrained with the initial weight values of the original dense model. 
{\color{black}
In JCW~\cite{zhao2022joint}, a supernet is used to evaluate subnet performance through joint channel and weight pruning. It searches for subnets considering tradeoff between latency and accuracy in the supernet by an evolutionary algorithm, and trains these subnets individually to obtain the final high-quality subnets. 
Our PSS-Net is very different from JCW:
1) The subnets of different sizes in JCW do not share weights because they are re-trained independently, while different subnets in our PSS-Net share weights. 
2) JCW uses the supernet as a subnet performance evaluator, while in PSS-Net the supernet also offers final weights of all subnets.
3) The search space in JCW includes entire channels and individual weights, while it is channels and resolutions in our PSS-Net.
4) JCW requires a heuristic algorithm in order to conduct subnet search, while by introducing a subnet pool of high-quality subnets, our PSS-Net eliminates the process of subnet search.
}
Besides, \cite{guo2020multi} and {\color{black}\cite{wang2022versatile}} also considered multi-dimension pruning including output channels, model depth, input resolution, \emph{etc}.
{\color{black}
Among them, VFS\cite{wang2022versatile} and our PSS-Net are similar in that they sample subnets under the constraints such as FLOPs and latency. The difference is that each subnet of VFS requires additional tuning therefore subnets do not share weights, while subnets in PSS-Net share weights since no fine-tuning is required therefore our PSS-Net saves more storage costs.
}

\textbf{Sample-Adaptive Networks}.
A sample-adaptive network adapts its structure according to the input image samples. Lin \emph{et al}.~\cite{lin2017runtime} modeled the adaption of network width as a Markov decision process, and measured the convolutional kernel importance conditioned on different samples. In~\cite{bejnordi2019batch} the network width is adjusted via a gating module conditioned on the current input. To this end, a ``batch-shaping'' is introduced to match the marginal aggregated posterior distribution over any intermediate features to a pre-specified prior distribution. Except for adjusting network width, another group realizes sample-adaptive network from the perspective of depth adaption. For instance, \cite{huang2017multi} proposed to early exit from the inference at shallow layers of the network through training multiple classifiers upon intermediate features. SkipNet~\cite{wang2018skipnet} skips convolutions via a gating network that maps the last-layer outputs to a binary decision to execute or bypass the current layer.

Traditional compact networks are fixed for deployment. A sample-adaptive network allocates more computational resources to complex images, and fewer to images that are easier for the task. As a core distinction, our PSS-Net produces multiple subnets and switches these subnets for deployment according to available hardware resources.

\section{Methodology}\label{sec:methodology}

\subsection{Preliminary}\label{sec:preliminary}
The proposed PSS-Net is an $L$-{\color{black}layer} supernet $\mathcal{N}_{\phi^{max}, \mathbf{W}}$ with its structure $\phi^{max}=(\{{o_i^{max}\}_{i=1}^{L}, r^{max}})$ and weights $\mathbf{W}=\{\mathbf{W}_i\}_{i=1}^L$, where $o_i^{max}$, $r^{max}$ and $\mathbf{W}_i$ are the number of output channels in the $i$-th layer, the supernet input resolution, and the weights of the $i$-th layer, respectively.
The supernet $\mathcal{N}_{\phi^{max}, \mathbf{W}}$ contains subnets of multiple sizes, and each subnet $\mathcal{N}_{\phi^t, \mathbf{W}^t}$ is allowed to use a different subnet structure $\phi^{t}=(\{{o_i^{t}\}_{i=1}^{L}, r^{t}})$ and subnet weights ${\mathbf{W}^t}=\{\mathbf{W}_i[1:o_i^{t}]\}_{i=1}^L$, similarly, where $o_i^{t}$ and $r^{t}$ are the number of output channels in the $i$-th layer and the subnet input resolution, respectively. $\mathbf{W}_i[1:o_i^{t}]$ means the subnet weights of the $i$-th layer are the weights of the first $o_i^t$ channels in $\mathbf{W}_i$. Therefore, any two subnets share partial weights of the supernet. 
Then we set separate lower bounds for each layer of the subnet structure $\{{o_i^{t}\}_{i=1}^{L}}$ and $r^{t}$ as $\{{o_i^{min}\}_{i=1}^{L}}$ and $r^{min}$, that is, the structure of the smallest subnet can be represented as $\phi^{min}=(\{{o_i^{min}\}_{i=1}^{L}}, r^{min})$.
Finally, we can represent the subnet structure space as $\mathcal{S}=\{\phi^1=(\{o_i^1\}_{i=1}^L, r^1), \phi^2=(\{o_i^2\}_{i=1}^L, r^2), ...\}$, where $o_i^{min} \leq o_i^t \leq o_i^{max}$, $r^{min} \leq r^t \leq r^{max}$.
In our practical implementation, for any subnet $\mathcal{N}_{\phi^t,\mathbf{W}^t}$, where the structure $\phi^t= \{ \{o^t_i \}_{i=1}^L, r^t \} \in \mathcal{S}$, we first sample $o^t_i$ from $\{o^{min}_i,~o^{min}_i+1,~o^{min}_i+2,~...,~o^{max}_i\}$ and then round to its nearest multiple of $8$ in order to obtain a hardware friendly channel number for the deployment on most modern devices. {\color{black}Specifically, in most hardware, setting the channel number to an integer multiple of 8, 16, or 32 aligns the matrix computation with the processor unit, allowing for faster matrix computation. For example, the GPU often has a wrap size of 32, while many mobile devices have a CPU with a wrap size of 8. Our method is designed to accommodate with more mobile devices, so we choose 8 as the channel divisor. The effect of different divisor sizes on the accuracy of the subnet will be discussed in Sec.\,\,\ref{sec:ablation}.}
Now $r^t$ can be sampled from the set of $\{r^{min},~r^{min}+8,~r^{min}+16,~...,~r^{max}\}$.

Given a set of resource constraints $\Theta = \{\theta^t\}_{t=1}^T$, the goal in this paper is to train a supernet $\mathcal{N}_{\phi^{max}, \mathbf{W}}$ such that a total of $T$ subnets $\{ \mathcal{N}_{\phi^t, \mathbf{W}^t} | \phi^t \in \mathcal{S}, \mathbf{W}^t \in \mathbf{W}\}_{t=1}^T$ conditioned on the resource constraints $\Theta$ can be directly extracted from this supernet. Therefore, we can now formally define the problem of our supernet training as:
\begin{equation}\label{eq:definition}
\begin{split}
    \{ (\phi^t)_{best} &\}_{t=1}^T, \{(\mathbf{W}^t)_{best}\}_{t=1}^T \\& = \underset{\{ \phi^t \}_{t=1}^T, \{\mathbf{W}^t\}_{t=1}^T}{\arg\max}\sum_{t=1}^T {\color{black}\text{Acc}}(\mathcal{N}_{\phi^t, \mathbf{W}^t}), \\
    \text{s.t.} \quad \phi^t \in & \mathcal{S}, \theta^t \in \Theta, \mathcal{C}(\mathcal{N}_{\phi^t, \mathbf{W}^t}) \in [\theta^t - \delta, \theta^t + \delta],
\end{split}
\end{equation}
where $\mathcal{C}(\mathcal{N}_{\phi^t, \mathbf{W}^t})$ calculates the resource consumption of the subnet $\mathcal{N}_{\phi^t, \mathbf{W}^t}$, $\delta$ is a bias tolerance between the actual resource consumption and the given resource constraint $\theta^t$, and ${\color{black}\text{Acc}}(\mathcal{N}_{\phi^t, \mathbf{W}^t})$ returns the accuracy of $\mathcal{N}_{\phi^t, \mathbf{W}^t}$ on the validation dataset. Note that, weights of each subnet are a subset of the supernet weights, \emph{i.e.}, $\mathbf{W}^t \in \mathbf{W}$. Thus, to update $\mathbf{W}^t$ is to update $\mathbf{W}$ in essence. For ease of the following presentation, we remove the superscript ``$t$'' in $\mathbf{W}^t$ and Eq.\,(\ref{eq:definition}) can be simplified as:
\begin{equation}\label{eq:sim_definition}
\begin{split}
    &\{ (\phi^t)_{best} \}_{t=1}^T, (\mathbf{W})_{best} = \underset{\{ \phi^t \}_{t=1}^T, \mathbf{W}}{\arg\max}\sum_{t=1}^T {\color{black}\text{Acc}}(\mathcal{N}_{\phi^t, \mathbf{W}}), \\
    & \textit{s.t.} \quad \phi^t \in \mathcal{S}, \theta^t \in \Theta,\ \mathcal{C}(\mathcal{N}_{\phi^t, \mathbf{W}}) \in [\theta^t - \delta, \theta^t + \delta].
\end{split}
\end{equation}

To train an $L$-layer supernet $\mathcal{N}_{\phi^t, \mathbf{W}}$, a three-step slimmable training paradigm is first proposed in~\cite{yu2019universally}, which consists of three steps: ``sandwich rule'', ``inplace distillation'', and ``recalibration of batch normalization layers''. More specifically, the ``sandwich rule'' is firstly adopted to select three subnet types of one maximal size (\emph{i.e.}, the supernet, $\mathcal{N}_{\phi^{max}, \mathbf{W}}$), several medium sizes and one minimal size (\emph{i.e.}, the minimal subnet $\mathcal{N}_{\phi^{min}, \mathbf{W}}$). Then, the ``inplace distillation'' is conducted between the supernet and other subnets. Finally, the parameters in the batch normalization layers of the different subnets are recalibrated after supernet training while keeping the convolutional layer parameters being frozen.
This paradigm has inspired many subsequent researchers to train resource-adaptive supernets or compact networks~\cite{yang2020mutualnet,cai2019once,lou2021dynamic,yu2020bignas,wang2021attentivenas}. 
The main difference between these approaches lies in the selection of the medium-sized subnets in the ``sandwich rule'' step. 
Traditional approaches~\cite{yu2019universally,yang2020mutualnet} obtain several medium-sized subnet using uniform subnet sampling, \emph{i.e.}, applying the same width multipliers to all layers of the supernet to obtain subnets of different sizes in each batch training. However, such a manner causes suboptimal performance due to the homogeneity of subnets.
The one-shot NAS studies~\cite{lou2021dynamic,yu2020bignas,wang2021attentivenas} randomly sample several medium-sized subnets from the pre-defined subnet structure space $\mathcal{S}$ to train the one-hot supernet. However, heavy cost is required to search for the target subnet when the available hardware resource constraints change from time to time.

\begin{algorithm}[t]
\caption{PSS-Net Training}\label{alg:PSS-Net}
\textbf{Input}: Supernet weight $\mathbf{W}$, supernet structure $\phi^{max}$, minimal subnet structure $\phi^{min}$, resource constraints $\Theta=\{ \theta^t \}_{t=1}^T$, subnet structure space $\mathcal{S}$.\\
\textbf{Output}: best subnet structures $\{ (\phi^t)_{best}\}_{t=1}^T$, Trained supernet weights $(\mathbf{W})_{best}$.

\begin{algorithmic}[1] 

\STATE Initialize subnet pools set: $\Pi=\{\pi^t=\{\}\}_{t=1}^T$.

{\scriptsize \# $e$: training epoch number, $b$: training batch number.}
\FOR{$e = 1, ..., E$} 
    \FOR{$b=1, ..., B$}
        \STATE $subnet\_list$ = $\{\phi^{max}\}$;
        \STATE Randomly sample $t$ from $\{1,2, ..., T\}$;
        \STATE $\phi^t, \pi^t$ = Alg.\,\ref{alg:sampling}($\theta^t, \pi^t, \mathcal{S}, e, E$);
        \STATE $subnet\_list$ = $subnet\_list$ $\cup\ \{\phi^t, \phi^{min}\}$;
        
        \STATE $\mathbf{W}$ = slimmable\_training($\mathbf{W}$, $subnet\_list$); \\
        {\scriptsize \# Details of slimmable\_training can be found in \ref{alg:slimmable}}
    \ENDFOR
\ENDFOR
\FOR{$t$ in $1,...,T$}
    \FOR{$j$ in $1,...,M$}
        \STATE Calibrate BN statistics of subnet $\mathcal{N}_{\phi^t_j, \mathbf{W}}$;
        \STATE Compute ${\color{black}\text{Acc}}(\mathcal{N}_{\phi^t_j, \mathbf{W}})$ on the validation set;
    \ENDFOR
     \STATE $(\phi^t)_{best}, (\mathbf{W})_{best} = \underset{(\phi^t)_j, \mathbf{W}}{\arg\max}{\color{black}\,\text{Acc}}(\mathcal{N}_{(\phi^t)_j, \mathbf{W}})$.
\ENDFOR
\STATE \textbf{return} $\{ (\phi^t)_{best}\}_{t=1}^T, (\mathbf{W})_{best}$.
\end{algorithmic}
\end{algorithm}

\begin{algorithm}[!ht]
\caption{Slimmable Training}
\label{alg:slimmable}
\textbf{Input}: Supernet weight $\mathbf{W}$, list for subnet structure $subnet\_list = \{\phi^{max}, \phi^{t}, \phi^{min}\}$.

\textbf{Output}: Supernet weight $\mathbf{W}$.

\begin{algorithmic}[1] 
\STATE Get current data batch $(x, y)$;
\STATE $\phi^{max}=subnet\_list[1]$;
\STATE Forward the maximum subnet as $\hat y=\mathcal{N}_{\phi^{max},\mathbf{W}}(x)$;
\STATE Obtain the cross-entropy loss as $loss=\text{CE}(\hat y,y)$;
\STATE Backward and accumulate loss as $loss.backward()$;
\STATE Detach the gradient of $\hat y$ as $\hat y=\hat y.detach()$;

\FOR{$\phi^t$ in $subnet\_list[2:]$}
    \STATE Forword subnet $y'=\mathcal{N}_{\phi^{t},\mathbf{W}}(x)$;
    \STATE Conduct knowledge distillation between $\mathcal{N}_{\phi^{max},\mathbf{W}}$ and $\mathcal{N}_{\phi^{t},\mathbf{W}}$ Compute KL loss using Kullback-Leibler loss as $loss=\text{KL}(y',\hat y)$;
    \STATE Backword and accumulate loss as $loss.backward()$;
\ENDFOR
\STATE Update supernet weight $\mathbf{W}$ as $optimizer.step()$;
\STATE \textbf{Return} $\mathbf{W}$.
\end{algorithmic}
\end{algorithm}

\subsection{PSS-Net Training}\label{sec:PSS-Net}
To solve the above problems, we propose prioritized subnet sampling to sample the medium-sized subnet for training our supernet, termed PSS-Net in this paper. 
Our PSS-Net training generally follows the three-step slimmable training paradigm as shown in Fig.\,\ref{fig:framework}. We first outline our PSS-Net training in Alg.\,\ref{alg:PSS-Net}, which is then detailed below.

\textbf{Subnet Pools (Line 1, Alg.\,\ref{alg:PSS-Net})}.
Different from existing methods, our PSS-Net training choose to maintain a set of subnet pools, denoted as $\Pi=\{\pi^t=\{\}\}_{t=1}^T$. The $t$-th subnet pool $\pi^t$ is used to collect the information of $M$ high-quality subnets conditioned on $\theta^t$ in Eq.\,(\ref{eq:sim_definition}).
The information of each subnet includes its network structure $\phi^t$ and performance metric $m^t$, which will be detailed in the next section.

\textbf{Slimmable Training (Lines 2 -- 10, Alg.\,\ref{alg:PSS-Net})}.
In each training batch, we select three subnet types of one maximal size (\emph{i.e.}, the supernet, $\mathcal{N}_{\phi^{max}, \mathbf{W}}$), one medium size and one minimal size (\emph{i.e.}, the minimal subnet $\mathcal{N}_{\phi^{min}, \mathbf{W}}$). The ``slimmable\_training'' in Line 8 of Alg.\,\ref{alg:PSS-Net} is the same as Line 3 -- Line 16 of Alg.\,\ref{alg:PSS-Net} in USNet~\cite{yu2019universally} and is elaborated 
in Alg.\,\ref{alg:slimmable}. 
To obtain a medium-sized subnet $\phi^t$, USNet simply applies a fixed width multiplier to uniformly reduce the width of the supernet, whereas with the aid of the introduced subnet pools, our PSS-Net gives priority to high-quality subnets, as outlined in Alg.\,\ref{alg:sampling} and detailed in the next section.

\textbf{BN Calibration (Lines 11 -- 17, Alg.\,\ref{alg:PSS-Net})}.
After the slimmable training, each subnet pool $\pi^t \in \Pi$ will be filled with these high-quality subnets conditioned on its corresponding resource constraint $\theta^t$. However, the statistics in the batch normalization (BN) layers of these high-quality subnets are not accurately estimated since the forward propagated subnets vary during training our PSS-Net. As a consequence, BN calibration is indispensable in order to evaluate the performance of each subnet. To this end, following~\cite{yu2019universally,yang2020mutualnet}, we feed each subnet with a small training set of $8,192$ images in this paper to update the BN statistics of each subnet in $\pi^t$. Then, the calibrated subnet with the best accuracy is regarded as the optimal subnet $\mathcal{N}_{(\phi^t)_{best},(\mathbf{W})_{best}}$ under the resource constraint of $\theta^t$. 
At the end of training, the subnet pool has collected a sufficient number of high-quality subnet structures for each resource constraint and the corresponding performance metrics detailed in the next section.
We only calibrate $k$ subnets within top-$k$ performance metrics, and then pick the one with the best accuracy from these chosen subnets.

\subsection{Prioritized Subnet Sampling}\label{sec:Prioritized}
We outline our prioritized subnet sampling in Alg.\,\ref{alg:sampling}, which serves as a core distinction of our work from the previous studies. More details are discussed in the following.

\textbf{Subnet Sampling (Lines 1 -- 7, Alg.\,\ref{alg:sampling})}.
As mentioned above, we intend to sample a medium-sized subnet $\phi^t$ to train our PSS-Net in each training batch. Instead of simply sampling subnets from the subnet structure space $\mathcal{S}$, we also conduct subnet sampling from the subnet pool $\pi^t$ given the resource constraint $\theta^t$.
To this end, we introduce a factor $p \in (0, 1)$ to indicate the probability of sampling from the subnet structure space $\mathcal{S}$.
At the beginning of PSS-Net training, we sample subnets from $\mathcal{S}$ until $\pi^t$ is full. In this situation, we expect $p = 1$. As the training of PSS-Net, more high-quality structures of subnets will be inserted into $\pi^t$, and we expect to give priority to sampling from $\pi^t$, which indicates a small value of $p$.
To this end, we formalize $p$ as:
\begin{equation}\label{p}
p = {p_{end}}^{\mathcal{I}(|\pi^t|< M)\cdot(e - 1)/E},
\end{equation}
where $p_{end}$ is a pre-given parameter indicating the probability of sampling 
from the subnet structure space $\mathcal{S}$ 
when PSS-Net training ends. In our implementation, we set $p_{end} = 0.01$. $\mathcal{I}(\cdot)$ is an indicator which returns 1 if the input is true, and 0 otherwise. The $e \in [1, E]$ and $E$ respectively denote the current training epoch and the total training epochs. As can be seen, starting from $1$, the probability of sampling from $\mathcal{S}$ decreases to $p_{end}$ gradually. Consequently, the priority will be entitled to the high-quality subnets.

\begin{algorithm}[!t]
\caption{Prioritized Subnet Sampling}
\label{alg:sampling}
\textbf{Input}: Subnet pool $\pi^t$ conditioned on resource constraint $\theta^t$, subnet structure space $\mathcal{S}$, current epoch $e$, and total epochs $E$. \\
\textbf{Output}: Subnet structure $\phi^t$, and subnet pool $\pi^t$

\begin{algorithmic}[1]
\STATE $p = {p_{end}}^{\mathcal{I}(|\pi^t|< M)\cdot(e - 1)/E}$;
\IF{$rand(0, 1)<p$}
    \STATE Sample a subnet structure $\phi^t \in \mathcal{S}$, s.t.$\ \mathcal{C}(\mathcal{N}_{\phi^t, \mathbf{W}}) \in [\theta^t-\delta, \theta^t+\delta]$ using Eq.\,(\ref{prob});
\ELSE
    \STATE $\eta = {\eta_{end}}^{\mathcal{I}(|\pi^t|< M)\cdot(e - 1)/E}$;
    \STATE Sample a subnet structure $\phi^t \in \pi^t$ using Eq.\,(\ref{pool_p});
\ENDIF
\IF{$\phi^t$ not in $\pi^t$ }
    \STATE $m^t = -batch\_loss$;
    \STATE Insert ($\phi^t, m^t$) into $\pi^t$;
    \IF{$|\pi^t| > M$}
        \STATE Remove $\big( (\phi^t)_M, (m^t)_M \big)$;
    \ENDIF
\ELSE
    \STATE $m^t = \lambda \cdot m^t + (1-\lambda)\cdot(-batch\_loss)$;
    \STATE Update $m^t$ of $\phi^t$ in $\pi^t$.
\ENDIF
\STATE \textbf{return} $\phi^t$, $\pi^t$.
\end{algorithmic}
\end{algorithm}

\textit{Case 1: Sampling from the subnet structure space $\mathcal{S}$ (Line 3 -- Line 4, Alg.\,\ref{alg:sampling})}.
To sample a subnet $\mathcal{N}_{\phi^t, \mathbf{W}}$ conditioned on $\theta^t$, one naive solution is to go through trial and error until an eligible subnet is found. However, this strategy brings heavy cost. Usually, thousands of sampling are inevitable each time according to our experimental observation.
Inspired by~\cite{wang2021attentivenas}, we resort to the probability distribution of subnets conditioned on the given resource constraint $\theta^t$, denoted as $\tau(\phi = \{o_i\}_{i=1}^L | \theta^t)$. However, the practical distribution is extremely complex to describe. While the Monte-Carlo approximation can be adopted, it is still intractable since multiple approximations have to be repeatedly constructed for each resource constraint $\theta^t \in \Theta$.
To overcome this, we first generate a large-scale set of subnet structures $\mathcal{G}$, and then compute the practical resource consumption of each structure $\phi \in \mathcal{G}$ as $\mathcal{C}(\mathcal{N}_{\phi, \mathbf{W}})$. Finally, we split $\mathcal{G}$ into $T$ non-overlapping subsets with $\mathcal{G}^1 \cup \mathcal{G}^2 \cup ... \cup \mathcal{G}^T = \mathcal{G}$. For each subnet structure $\phi^t \in \mathcal{G}^t$, its resource consumption satisfies that $\mathcal{C}(\mathcal{N}_{\phi^t, \mathbf{W}}) \in [\theta^t - \sigma, \theta^t + \sigma]$.
To model $\tau(\phi^t = \{o_i^t\}_{i=1}^L | \theta^t)$, we directly estimate the marginal probability of $\tau(o_i^t | \theta^t)$ as:
\begin{equation}\label{prob}
\tau(o_i^t | \theta^t) \approx \frac{\sum_{\phi^t = \{o^t_k\}_{k=1}^L \in \mathcal{G}^t}\mathcal{I}(o_k^t = o_i^t)}{| \mathcal{G}^t |}.
\end{equation}

With Eq.\,(\ref{prob}), the sampling times to find the target subnet reduces from $10^6$ to $10^1$ approximately according to our observation, which greatly improves the sampling efficiency.

\textit{Case 2: Sampling from the subnet pool $\pi^t$ (Line 5 -- Line 7, Alg.\,\ref{alg:sampling}).}
For $\pi^t = \{ \big( (\phi^t)_j, (m^t)_j ) \big) \}_{j=1}^M$, 
to obtain the best subnet conditioned on $\theta^t$ in Eq.\,(\ref{eq:sim_definition}), we resort to preserving the best subnet structure in $\pi^t$ for deployment, \emph{i.e.}, 
\begin{equation}
(\phi^t)_{best} = \underset{(\phi^t)_j}{\arg\max}\, {\color{black}\text{Acc}}(\mathcal{N}_{(\phi^t)_j,\mathbf{W}}),
\end{equation}
after our PSS-Net training. Thus, it is natural to prioritize the subnet structure with the best performance $m^t$ in $\pi^t$ to train our PSS-Net. To this end, we perform weighted sampling upon $\pi^t$ so that better subnets are assigned with higher probabilities. Thus, we can obtain a distribution set $\{ (q^t)_j \}_{j=1}^M$ with $\sum_{j=1}^M(q^t)_j = 1$ where $(q^t)_j$ denotes the probability of sampling $(\phi^t)_j$ to train PSS-Net, defined as:
\begin{equation}\label{pool_p}
    (q^t)_j = \frac{\exp{\big((m^t)_j / \eta}\big)}{\sum_{i=1}^M\exp{\big((m^t)_i / \eta\big)}},
\end{equation}
where $\eta$ is a temperature to control the smoothness of the distribution. In the early training stage, the subnets are not well trained, and thus $m^t$ is not a reliable measure to reflect the actual performance of each subnet. At this time, we expect each subnet to be sampled equally. As training goes, $m^t$ will become stable and we expect the distribution to be a one-hot form. Similar to Eq.\,(\ref{p}), we achieve this by setting $\eta = {\eta_{end}}^{\mathcal{I}(|\pi^t|< M)\cdot(e - 1)/E}$, where $\eta_{end} = 0.01$.

\textbf{Subnet Pool Manipulation (Lines 8 -- 16, Alg.\,\ref{alg:sampling})}.
We continuously collect the information tuple $(\phi^t, m^t)$ until the pool $\pi^t$ is full. When receiving a $(\phi^t, m^t)$, we insert it into $\pi^t$ such that the information tuples in $\pi^t$ are arranged in a decreasing order according to the performance $m^t$. In the case of full pool, we can obtain $\pi^t = \{ \big( (\phi^t)_j, (m^t)_j ) \big) \}_{j=1}^M$ with $(m^t)_{j+1} \le (m^t)_j$ and $(\phi^t)_j = \{(o^t_i)_j\}_{i=1}^L$. If one more tuple received, we will remove the worst performing tuple, \emph{i.e.}, $\big((\phi^t)_M, (m^t)_M\big)$, before inserting this tuple.

To model $m^t$ of $\phi^t$, the best way is to define it as the accuracy of the subnet $\mathcal{N}_{\phi^t, \mathbf{W}}$. However, due to the large subnet structure space, it is almost impossible to execute a complete training for every subnet. Inspired by~\cite{wang2021attentivenas,berman2020aows}, we define $(m^t)_j = -batch\_loss$ if $\phi^t$ is inserted into $\pi^t$ for the first time. If $\phi^t$ is sampled from $\pi^t$ for updating PSS-Net, we update $m^t$ using moving average as $m^t = \lambda \cdot m^t + (1-\lambda)\cdot(-batch\_loss)$ after the current training batch. Here, we define $batch\_loss$ as the training loss in the current batch.

\begin{figure}[!t]
\centering
\includegraphics[width=0.8\columnwidth]{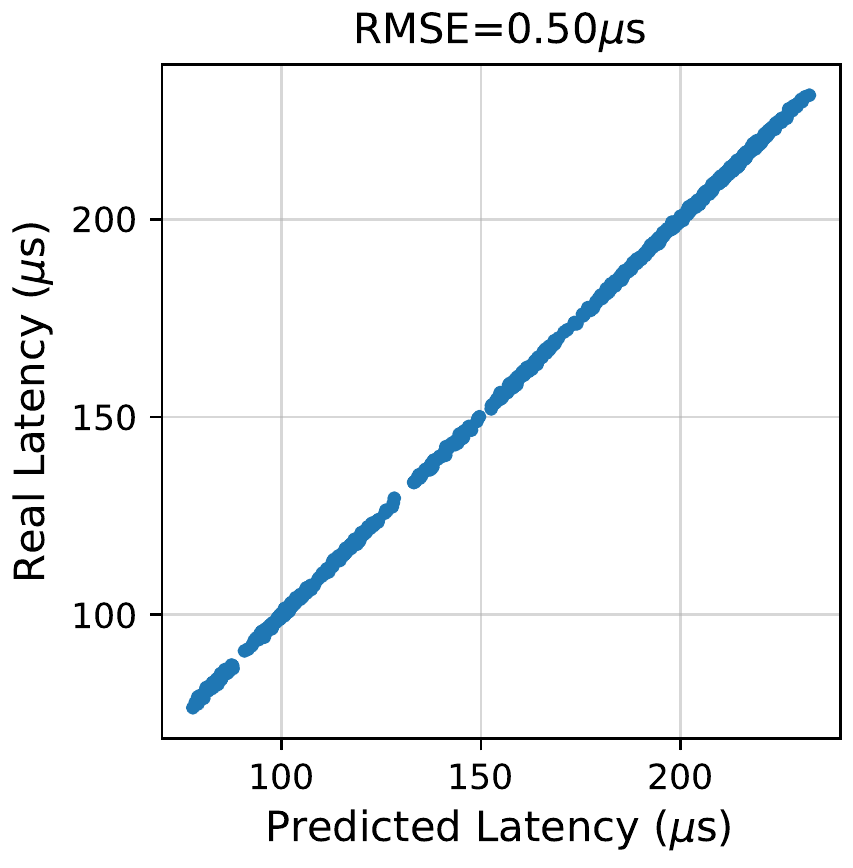}
\vspace{-1.0em}
\caption{Predicted latency \emph{v.s.} real latency. We sample 1,000 subnet structures using MobileNet-V1 as the supernet. We obtain a very small root mean square error (RMSE) of 0.5$\mu$s between real latency and predicted latency, indicating an accurate prediction of our lookup tables.}
\label{fig:lat}
\vspace{-1.0em}
\end{figure}

\begin{table*}[!t]
\caption{
Performance comparison with the resource-adaptive methods USNet~\cite{yu2019universally} and MutualNet~\cite{yang2020mutualnet}. 
PSS-Net$_\text{F}$, PSS-Net$_\text{G}$, and PSS-Net$_\text{C}$ denote PSS-Net trained under individual constraint of FLOPs, GPU latency, and CPU latency.
PSS-Net$_\text{M}$ denotes PSS-Net trained under all of the three constraints.
{\color{black}Using MobileNet-V1, MobileNet-V2 and ResNet-50 as backbone networks.}
The ``-'' denotes that the corresponding method does not have a subnet that satisfies the required constraint.
}
\vspace{-1.0em}
\centering
\setlength{\tabcolsep}{2.5mm}
{
{\color{black}
\begin{tabular}{c|ccc|ccc|ccc} 
\toprule
\multirow{13}{*}{MobileNet-V1} & Method & FLOPs~(M) & Acc@1~(\%) & Method & GPU~($\mu$s) & Acc@1~(\%) & Method & CPU~(ms) & Acc@1~(\%) \\ 
\cline{2-10}
 & USNet & 569 & 71.8 & USNet & 238 & 71.8 & USNet & 36 & 71.8 \\
 & MutalNet & 569 & 72.4 & MutalNet & 238 & 72.4 & MutalNet & 37 & 72.4 \\
 & PSS-Net$_\text{F}$ & \textbf{555} & \textbf{74.2} & PSS-Net$_\text{G}$ & \textbf{238} & \textbf{74.4} & PSS-Net$_\text{C}$ & \textbf{32} & \textbf{74.2} \\
 & PSS-Net$_\text{M}$ & \textbf{562} & \textbf{74.5} & PSS-Net$_\text{M}$ & \textbf{236} & \textbf{74.5} & PSS-Net$_\text{M}$ & \textbf{30} & \textbf{74.5} \\ 
\cline{2-10}
 & USNet & 306 & 69.1 & USNet & 138 & 66.3 & USNet & 22 & 67.8 \\
 & MutalNet & 309 & 69.7 & MutalNet & 132 & 68.4 & MutalNet & 21 & 68.4 \\
 & PSS-Net$_\text{F}$ & \textbf{306} & \textbf{72.3} & PSS-Net$_\text{G}$ & \textbf{128} & \textbf{71.9} & PSS-Net$_\text{C}$ & \textbf{21} & \textbf{72.8} \\
 & PSS-Net$_\text{M}$ & \textbf{309} & \textbf{72.5} & PSS-Net$_\text{M}$ & \textbf{131} & \textbf{71.8} & PSS-Net$_\text{M}$ & \textbf{21} & \textbf{72.5} \\ 
\cline{2-10}
 & USNet & 114 & 61.9 & USNet & 76 & 57.3 & USNet & 13 & 61.0 \\
 & MutalNet & 111 & 63.5 & MutalNet & 80 & 63.5 & MutalNet & 13 & 63.5 \\
 & PSS-Net$_\text{F}$ & \textbf{107} & \textbf{67.9} & PSS-Net$_\text{G}$ & \textbf{76} & \textbf{68.1} & PSS-Net$_\text{C}$ & \textbf{13} & \textbf{68.3} \\
 & PSS-Net$_\text{M}$ & \textbf{107} & \textbf{68.0} & PSS-Net$_\text{M}$ & \textbf{71} & \textbf{68.1} & PSS-Net$_\text{M}$ & \textbf{13} & \textbf{68.6} \\ 
\hline\hline
\multirow{13}{*}{MobileNet-V2} & Method & FLOPs (M) & Acc@1 (\%) & Method & GPU ($\mu$s) & Acc@1 (\%) & Method & CPU (ms) & Acc@1 (\%) \\ 
\cline{2-10}
 & USNet & 301 & 71.5 & USNet & 269 & 71.5 & USNet & 41 & 71.5 \\
 & MutalNet & 301 & 72.9 & MutalNet & 275 & 72.9 & MutalNet & 41 & 72.9 \\
 & PSS-Net$_\text{F}$ & \textbf{301} & \textbf{73.4} & PSS-Net$_\text{G}$ & \textbf{260} & \textbf{73.4} & PSS-Net$_\text{C}$ & \textbf{30} & \textbf{73.2} \\
 & PSS-Net$_\text{M}$ & \textbf{295} & \textbf{73.4} & PSS-Net$_\text{M}$ & \textbf{260} & \textbf{73.4} & PSS-Net$_\text{M}$ & \textbf{33} & \textbf{73.4} \\ 
\cline{2-10}
 & USNet & 151 & 67.6 & USNet & 229 & 62.3 & USNet & - & - \\
 & MutalNet & 154 & 70.1 & MutalNet & 207 & 71.0 & MutalNet & 25 & 71.0 \\
 & PSS-Net$_\text{F}$ & \textbf{154} & \textbf{70.7} & PSS-Net$_\text{G}$ & \textbf{197} & \textbf{72.5} & PSS-Net$_\text{C}$ & \textbf{23} & \textbf{72.4} \\
 & PSS-Net$_\text{M}$ & \textbf{154} & \textbf{70.7} & PSS-Net$_\text{M}$ & \textbf{197} & \textbf{72.4} & PSS-Net$_\text{M}$ & \textbf{26} & \textbf{72.4} \\ 
\cline{2-10}
 & USNet & 88 & 62.2 & USNet & - & - & USNet & - & - \\
 & MutalNet & 84 & 65.8 & MutalNet & 105 & 64.2 & MutalNet & 17 & 64.2 \\
 & PSS-Net$_\text{F}$ & \textbf{84} & \textbf{66.8} & PSS-Net$_\text{G}$ & \textbf{101} & \textbf{66.5} & PSS-Net$_\text{C}$ & \textbf{17} & \textbf{69.0} \\
 & PSS-Net$_\text{M}$ & \textbf{83} & \textbf{66.7} & PSS-Net$_\text{M}$ & \textbf{102} & \textbf{66.7} & PSS-Net$_\text{M}$ & \textbf{18} & \textbf{68.2} \\ 
\hline\hline
\multirow{13}{*}{ResNet-50} & Method & FLOPs (G) & Acc@1 (\%) & Method & GPU ($\mu$s) & Acc@1 (\%) & Method & CPU (ms) & Acc@1 (\%) \\ 
\cline{2-10}
 & USNet & 4.1 & 76.2 & USNet & 911 & 76.2 & USNet & 109 & 76.2 \\
 & MutalNet & 4.1 & 76.6 & MutalNet & 911 & 76.6 & MutalNet & 110 & 76.6 \\
 & PSS-Net$_\text{F}$ & \textbf{3.5} & \textbf{77.3} & PSS-Net$_\text{G}$ & \textbf{858} & \textbf{77.2} & PSS-Net$_\text{C}$ & 104 & 77.2 \\
 & PSS-Net$_\text{M}$ & \textbf{3.5} & \textbf{77.4} & PSS-Net$_\text{M}$ & \textbf{864} & \textbf{77.4} & PSS-Net$_\text{M}$ & 109 & 77.5 \\ 
\cline{2-10}
 & USNet & 2.0 & 74.5 & USNet & 448 & 72.7 & USNet & 87 & 73.9 \\
 & MutalNet & 2.0 & 75.3 & MutalNet & 448 & 74.8 & MutalNet & 85 & 75.8 \\
 & PSS-Net$_\text{F}$ & \textbf{2.0} & \textbf{76.2} & PSS-Net$_\text{G}$ & \textbf{468} & \textbf{75.6} & PSS-Net$_\text{C}$ & \textbf{85} & \textbf{76.8} \\
 & PSS-Net$_\text{M}$ & \textbf{2.0} & \textbf{76.3} & PSS-Net$_\text{M}$ & \textbf{430} & \textbf{75.8} & PSS-Net$_\text{M}$ & \textbf{85} & \textbf{76.9} \\ 
\cline{2-10}
 & USNet & 1.1 & 72.4 & USNet & - & - & USNet & 60 & 71.7 \\
 & MutalNet & 1.0 & 73.0 & MutalNet & 209 & 72.5 & MutalNet & 61 & 72.5 \\
 & PSS-Net$_\text{F}$ & \textbf{0.8} & \textbf{73.5} & PSS-Net$_\text{G}$ & \textbf{210} & \textbf{73.5} & PSS-Net$_\text{C}$ & \textbf{58} & \textbf{73.4} \\
 & PSS-Net$_\text{M}$ & \textbf{0.8} & \textbf{73.8} & PSS-Net$_\text{M}$ & \textbf{211} & \textbf{73.8} & PSS-Net$_\text{M}$ & \textbf{60} & \textbf{73.8} \\
\bottomrule
\end{tabular}
}
}
\label{tab:main}
\end{table*}

\section{Experiments}\label{experiment}
\subsection{Experimental Settings}
We use the light-weight MobileNet-V1~\cite{howard2017mobilenets} and MobileNet-V2~\cite{sandler2018mobilenetv2} {\color{black}as well as heavy ResNet-50~\cite{he2016deep}} as the backbones of PSS-Net and compared methods, and then conduct {\color{black}classification} experiments on ImageNet~\cite{deng2009imagenet}. 
We set $o_i^{max}$ to be the same as the channel number of the $i$-th layer in the backbone network, and $r^{max}$ is 224. Then we define $o_i^{min}=0.75 \cdot o_i^{max}$, $r^{min}=128$. After that we can obtain the supernet $\mathcal{N}_{\phi^{max}, \mathbf{W}}$, and the subnet space $\mathcal{S}$.
We perform three resource constraints in this paper, including FLOPs, 
GPU and CPU latencies.
The resource constraint set $\Theta$ ranges from $\mathcal{C}(\mathcal{N}_{\phi^{min}, \mathbf{W}})$ to $\mathcal{C}(\mathcal{N}_{\phi^{max}, \mathbf{W}})$ with steps of 10M, 10$\mu$s and 1ms for FLOPs, GPU latency and CPU latency, respectively.
The subnet pool size $M$ is set to $50$. 

We construct lookup tables of CPU and GPU latencies for a quick calculation of resource consumption $\mathcal{C}(\cdot)$ during PSS-Net training. 
The details of the lookup tables and more ablations are provided in Sec.\,\,\ref{sec:lut}.
We use PyTorch as our training framework. The initial learning rate is set to 0.3 which is then adjusted by cosine annealing. SGD optimizer is adopted with a weight decay of 1e-4. We train our PSS-Net for a total of 250 epochs with a batch size of 1024.

\subsection{Latency Lookup Table}\label{sec:lut}

We respectively measure the CPU latency and GPU latency on a single core of \emph{Intel Xeon Platinum 8268} CPU and on a single \emph{NVIDIA Tesla V100} GPU 
with a batch size of 1 and 128.
To realize a quick calculation of resource consumption $\mathcal{C}(\cdot)$, we construct lookup tables of CPU and GPU latencies for our PSS-Net training. 
To this end, the most naive approach is to pre-compute the practical consumption on the target platform for all subnet structures $\phi^t \in \mathcal{S}$. However, the large volume of $\mathcal{S}$ makes it intractable. 
Instead, we choose to compute the latency of each subnet block in $\mathcal{S}$ and build latency tables for these blocks. During PSS-Net training, the latency of a given subnet structure is predicted by summing up the latencies of the blocks making up this subnet. In Fig.\,\ref{fig:lat}, we randomly sample 1,000 subnet structures using MobileNet-V1 as the supernet, and show the actual GPU latencies of these subnets \emph{v.s.} their block summation from the lookup tables. In Fig.\,\ref{fig:lat}, the predicted latency is very close to the real latency, which indicates that our lookup tables are very accurate and reliable.

\subsection{Experimental Results}
In this subsection, we compare the top-1 accuracy of subnets obtained by different methods, including the resource-adaptive methods and the compact network methods under similar resource constraints.

\textbf{Comparison with Resource-Adaptive Methods}.
We first compare our PSS-Net with the resource-adaptive studies~\cite{yu2019universally,yang2020mutualnet}. 
we train four types of PSS-Net constrained by different resources including: 
(1) PSS-Net$_\text{F}$ is trained under only FLOPs constraint;
(2) PSS-Net$_\text{C}$ is trained under only CPU latency constraint;
(3) PSS-Net$_\text{G}$ is trained under only GPU latency constraint;
(4) PSS-Net$_\text{M}$ is trained under multiple constraints of FLOPs, 
CPU latency and GPU latency.

In Table\,\ref{tab:main}, we compare the accuracies of the subnets resulting from different methods under similar FLOPs, GPU latency and CPU latency.
First, our PSS-Net provides subnets with better performance when compared to MutualNet~\cite{yang2020mutualnet} and USNet~\cite{yu2019universally} under similar or less resource consumption on FLOPs, GPU latency and CPU latency. 
Second, USNet using MobileNet-V2 as its supernet backbone fails to produce subnets with low GPU/CPU latency.
Third, though PSS-Net$_\text{M}$ needs to accommodate more resource constraints, its performance is still comparable or even better than PSS-Net$_\text{F}$, PSS-Net$_\text{G}$ and PSS-Net$_\text{C}$ which are dedicated to a single type of resource constraint.
Fourth, PSS-Net well improves the performance of backbone networks, for instance, PSS-Net$_\text{M}$ well increases the accuracy of MobileNet-V1/V2 from 70.9\%/71.8\% to 74.5\%/73.4\% with FLOPs reduction.

\begin{figure*}[!t]
\centering
\includegraphics[width=0.85\textwidth]{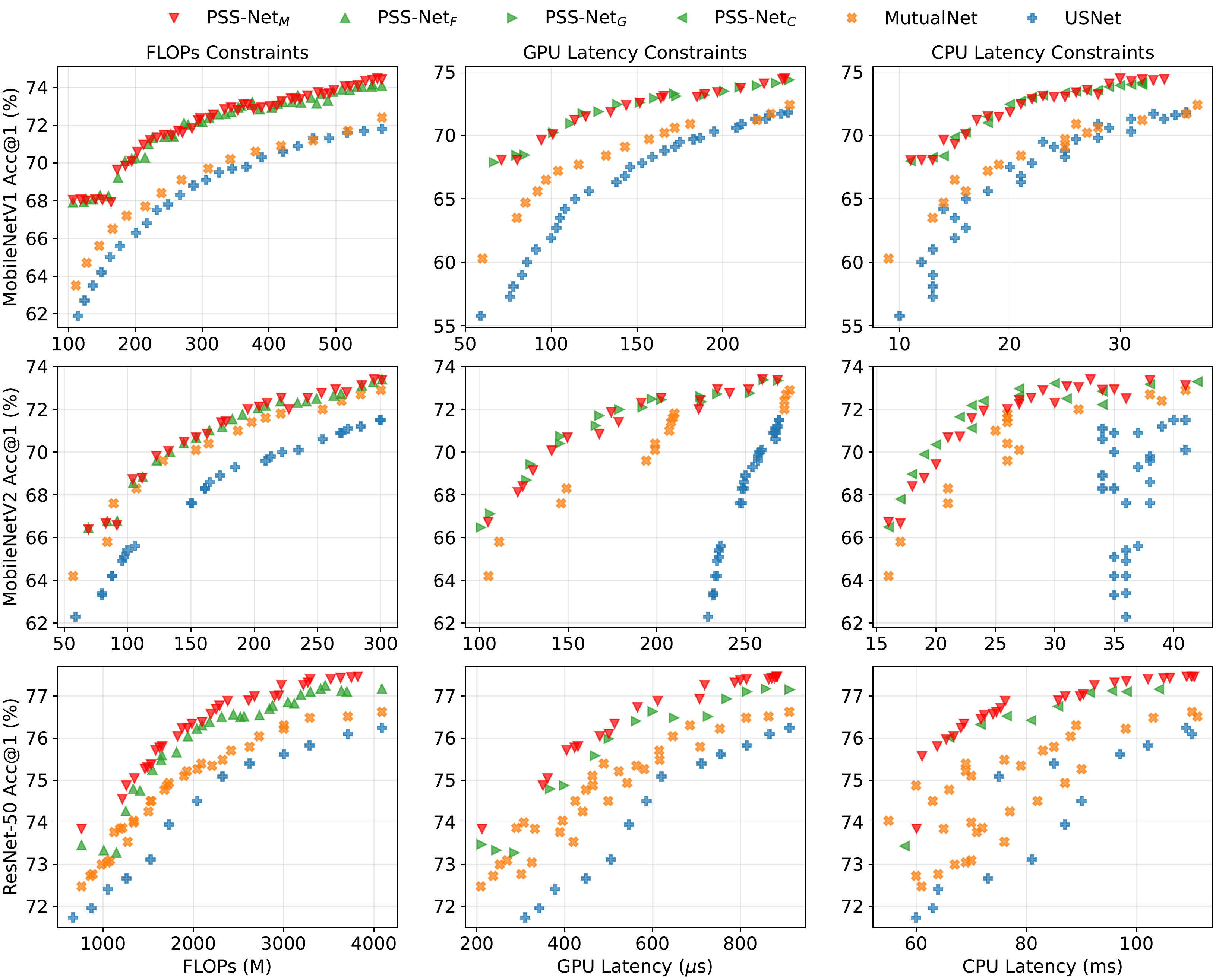}
\vspace{-1.5em}
\caption{MobileNet-V1, MobileNet-V2 {\color{black}and ResNet-50} experiments under single and multiple types of resource constraints.
PSS-Net$_\text{F}$, PSS-Net$_\text{G}$, and PSS-Net$_\text{C}$ denote PSS-Net trained under individual constraint of FLOPs, GPU latency, and CPU latency.
PSS-Net$_\text{M}$ denotes PSS-Net trained under all of the three constraints.
}
\label{fig:main}
\vspace{-1em}
\end{figure*}

To better demonstrate the advantage of our PSS-Net, we further plot the resource-accuracy curves of different methods in Fig.\,\ref{fig:main}.
We can observe that our PSS-Net consistently outperforms USNet and MutualNet by a large margin in all subfigures. Under similar resource consumption, our PSS-Net brings subnets with higher accuracy.
In addition, similar to Table\,\ref{tab:main}, USNet using MobileNet-V2 as the supernet backbone only produces subnets with high GPU/CPU latency. Specifically, the GPU latency is over 225$\mu$s and the CPU latency is over 30ms. Such a limitation greatly barricades the deployment of USNet, particularly when the available resources are in short supply. The reasons are attributed to two points. First, the input image resolution of USNet is fixed to 224, which is a large size and thus results in heavy GPU/CPU latency. Second, USNet adopts uniform subnet sampling, which ignores the diversity of layer-wise width. The resulting subnets are usually hardware-unfriendly. Our PSS-Net searches for the optimal image resolution for each subnet and the sampling strategy is non-uniform, which brings about subnets with low latency in comparison with existing methods. In addition to that, the proposed prioritized subnet sampling makes sure that high-performing subnets are trained sufficiently, which in turn entitles high performance of these subnets.

\begin{table*}[!t]
\centering
\caption{
Performance comparison with compact model methods including model pruning (${\color{blue}\blacktriangledown}$) and NAS (${\color{red}\bigstar}$).
The model pruning and PSS-Net methods in {\color{black}the left, middle and right} columns use MobileNet-V1, MobileNet-V2 {\color{black}and ResNet-50} as the supernet networks, respectively.
}
\label{tab:pruning}
\vspace{-1.0em}
\setlength{\tabcolsep}{2mm}
{
\begin{tabular}{ccc|ccc|ccc} 
\hline
\multicolumn{3}{c|}{MobileNet-V1} & \multicolumn{3}{c|}{MobileNet-V2} & \multicolumn{3}{c}{{ResNet-50}} \\ 
\hline
Method & FLOPs (M) & Acc@1 (\%) & Method & FLOPs (M) & Acc@1 (\%) & {Method} & {FLOPs (G)} & {Acc@1 (\%)} \\ 
\hline
Baseline & 569 & 70.9 & Baseline & 301 & 71.8 & {Baseline} & {4.1} & {76.15} \\
PNAS$\color{red}^\bigstar$ & 588 & 74.2 & NASNet$\color{red}^\bigstar$ & 488 & 72.8 & {MetaPruning$\color{blue}^\blacktriangledown$} & {3.0} & {76.2} \\
DARTS$\color{red}^\bigstar$ & 574 & 73.3 & MetaPruning$\color{blue}^\blacktriangledown$ & 291 & 72.7 & {PFS$\color{blue}^\blacktriangledown$} & {3.0} & {76.7} \\
NASNet$\color{red}^\bigstar$ & 564 & 74.0 & \textbf{PSS-Net$_\text{M}$} & \textbf{295} & \textbf{73.4} & \textbf{{PSS-Net$_\text{F}$}} & \textbf{{3.0}} & \textbf{{76.9}} \\
PFS$\color{blue}^\blacktriangledown$ & 567 & 71.6 & \textbf{PSS-Net$_\text{F}$} & \textbf{294} & \textbf{73.3} & \textbf{{PSS-Net$_\text{M}$}} & \textbf{{3.0}} & \textbf{{77.3}} \\
\textbf{PSS-Net$_\text{F}$} & \textbf{555} & \textbf{74.2} & DPFPS$\color{blue}^\blacktriangledown$ & 226 & 71.1 & {ThiNet$\color{blue}^\blacktriangledown$} & {2.6} & {74.0} \\
\textbf{PSS-Net$_\text{M}$} & \textbf{562} & \textbf{74.5} & ChamNet$\color{red}^\bigstar$ & 212 & 71.6 & {ABCPruner$\color{blue}^\blacktriangledown$} & {2.56} & {74.8} \\
MetaPruning$\color{blue}^\blacktriangledown$ & 324 & 70.9 & {VFS$\color{blue}^\blacktriangledown$} & {212} & {70.6} & {HRank$\color{blue}^\blacktriangledown$} & {2.3} & {75.0} \\
MDP$\color{blue}^\blacktriangledown$ & 309 & 71.2 & CC$\color{blue}^\blacktriangledown$ & 215 & 70.9 & {MetaPuning$\color{blue}^\blacktriangledown$} & {2.0} & {75.4} \\
\textbf{PSS-Net$_\text{F}$} & \textbf{310} & \textbf{72.4} & AMC$\color{blue}^\blacktriangledown$ & 211 & 70.8 & {PFS$\color{blue}^\blacktriangledown$} & {2.0} & {75.6} \\
\textbf{PSS-Net$_\text{M}$} & \textbf{309} & \textbf{72.5} & DMCP$\color{blue}^\blacktriangledown$ & 211 & 72.2 & {VFS$\color{blue}^\blacktriangledown$} & {2.0} & {75.3} \\
PFS$\color{blue}^\blacktriangledown$ & 286 & 70.7 & PFS$\color{blue}^\blacktriangledown$ & 210 & 70.9 & {FPGM$\color{blue}^\blacktriangledown$} & {1.9} & {74.8} \\
AMC$\color{blue}^\blacktriangledown$ & 285 & 70.5 & MDP$\color{blue}^\blacktriangledown$ & 206 & 71.4 & \textbf{{PSS-Net$_\text{F}$}} & \textbf{{1.9}} & \textbf{{76.0}} \\
NetAdapt$\color{blue}^\blacktriangledown$ & 284 & 69.1 & \textbf{PSS-Net$_\text{M}$} & \textbf{204} & \textbf{72.1} & \textbf{{PSS-Net$_\text{M}$}} & \textbf{{1.9}} & \textbf{{76.2}} \\
EagleEye$\color{blue}^\blacktriangledown$ & 284 & 70.9 & \textbf{PSS-Net$_\text{F}$} & \textbf{203} & \textbf{72.1} & {ThiNet$\color{blue}^\blacktriangledown$} & {1.16} & {68.2} \\
MetaPruning$\color{blue}^\blacktriangledown$ & 281 & 70.6 & CPLI$\color{blue}^\blacktriangledown$ & 166 & 67.4 & {CURL$\color{blue}^\blacktriangledown$} & {1.10} & {73.4} \\
\textbf{PSS-Net$_\text{M}$} & \textbf{278} & \textbf{71.8} & DMC$\color{blue}^\blacktriangledown$ & 163 & 68.4 & {MetaPuning$\color{blue}^\blacktriangledown$} & {1.00} & {73.4} \\
\textbf{PSS-Net$_\text{F}$} & \textbf{273} & \textbf{72.1} & \textbf{PSS-Net$_\text{F}$} & \textbf{163} & \textbf{70.9} & {ABCPruner$\color{blue}^\blacktriangledown$} & {0.94} & {70.3} \\
PFS$\color{blue}^\blacktriangledown$ & 150 & 65.5 & \textbf{PSS-Net$_\text{M}$} & \textbf{163} & \textbf{70.9} & {PFS$\color{blue}^\blacktriangledown$} & {1.00} & {72.8} \\
AutoSlim$\color{blue}^\blacktriangledown$ & 150 & 67.9 & MetaPruning$\color{blue}^\blacktriangledown$ & 140 & 68.2 & {HRank$\color{blue}^\blacktriangledown$} & {0.98} & {69.1} \\
MetaPruning$\color{blue}^\blacktriangledown$ & 149 & 66.1 & MDP$\color{blue}^\blacktriangledown$ & 139 & 68.8 & {VFS$\color{blue}^\blacktriangledown$} & {0.98} & {73.6} \\
\textbf{PSS-Net$_\text{F}$} & \textbf{126} & \textbf{68.1} & \textbf{PSS-Net$_\text{F}$} & \textbf{134} & \textbf{70.0} & \textbf{{PSS-Net$_\text{F}$}} & \textbf{{0.76}} & \textbf{{73.5}} \\
\textbf{PSS-Net$_\text{M}$} & \textbf{118} & \textbf{68.1} & \textbf{PSS-Net$_\text{M}$} & \textbf{133} & \textbf{70.0} & \textbf{{PSS-Net$_\text{M}$}} & \textbf{{0.76}} & \textbf{{73.8}} \\
\hline
\end{tabular}
}
\end{table*}

\textbf{Comparison with Compact Models}.
We compare PSS-Net with some compact models including model pruning and network architecture search (NAS) methods. 
Model pruning methods include {\color{black}VFS~\cite{wang2022versatile},} PFS~\cite{wang2020pruning}, MetaPruning~\cite{liu2019metapruning}, MDP~\cite{liu2020joint}, AMC~\cite{he2018amc}, NetAdapt~\cite{yang2018netadapt}, EagleEye~\cite{li2020eagleeye}, AutoSlim~\cite{yu2019autoslim}, DPFPS~\cite{ruan2021dpfps}, CC~\cite{li2021towards}, DMCP~\cite{guo2020dmcp}, CPLI~\cite{guo2020channel}, ThiNet~\cite{luo2017thinet}, ABCPruner~\cite{lin2020channel}, HRank~\cite{lin2020hrank}, FPGM~\cite{he2019filter}, CURL~\cite{luo2020neural} and DMC~\cite{gao2020discrete}.
NAS methods include DARTS~\cite{liu2019darts}, PNAS~\cite{liu2018progressive}, NASNet~\cite{zoph2018learning} and ChamNet~\cite{dai2019chamnet}.
The subnet search space of PSS-Net is limited to the size of the backbone networks, while the subnet search space of most NAS methods is much larger than that of the backbone networks. Therefore, for a fair comparison, we mainly compare PSS-Net with the model pruning methods.
We compare the accuracy under similar FLOPs consumption since the FLOPs metric is more often adopted in the compact model research.
To that effect, we train our supernet PSS-Net$_\text{M}$ and PSS-Net$_\text{F}$ using MobileNet-V1/V2 and {\color{black}ResNet-50} as backbones respectively, and then extract the best subnets from PSS-Net$_\text{M}$ and PSS-Net$_\text{F}$, whose FLOPs consumption is similar to the compact models for a fair comparison.
Table\,\ref{tab:pruning} displays the performance comparison. As can be seen, our PSS-Net$_\text{M}$ results in high-capacity subnets which not only surpass the compact model methods significantly, but also are better than the original MobileNet-V1/V2 {\color{black}and ResNet-50 sometimes}. Taking MobileNet-V1 as an example, our PSS-Net$\text{M}$ provides a subnet that reduces the FLOPs of MobileNet-V1 from 569M to 278M, and meanwhile increases the accuracy from 70.6\% to 71.8\%. Moreover, these compact methods produce static models for deployment while our PSS-Net$_\text{M}$ offers switchable subnets to adapt the change of available resources.

\begin{table}
\centering
{\color{black}
\caption{
MobileNet-V1 and MobileNet-V2 TFLite latency measured on ARM A72 single core CPU.
}
\label{cpu}
\vspace{-1.0em}
}
{\color{black}
\setlength{\tabcolsep}{1.3mm}
{
\begin{tabular}{c|cccc} 
\hline
Model & Method & \begin{tabular}[c]{@{}c@{}}Share\\Weight ?\end{tabular} & \begin{tabular}[c]{@{}c@{}}Latency\\(ms)\end{tabular} & Acc@1 (\%) \\ 
\hline
\multirow{13}{*}{MobileNet-V1} & Uniform 1.0$\times$ & N & 167 & 70.9 \\
 & Uniform 0.75$\times$ & N & 102 & 68.4 \\
 & USNet & Y & 102 & 69.5 \\
 & AutoSlim & N & 99 & 71.5 \\
 & AMC & N & 94 & 70.7 \\
 & JCW & N & 69 & 71.4 \\
 & \textbf{PSS-Net} & \textbf{Y} & \textbf{67} & \textbf{71.7} \\
 & FSC & N & 61 & 68.4 \\
 & AutoSlim & N & 55 & 67.9 \\
 & JCW & N & 54 & 70.3 \\
 & \textbf{PSS-Net} & \textbf{Y} & \textbf{54} & \textbf{70.6} \\
 & Uniform 0.5$\times$ & N & 53 & 64.4 \\
 & USNet & Y & 53 & 64.2 \\ 
\hline
\multirow{16}{*}{MobileNet-V2} & Uniform 1.0$\times$&N & 114 & 71.8 \\
 & APS & N & 110 & 72.8 \\
 & FSC & N & 89 & 72.0 \\
 & DMCP & N & 83 & 72.4 \\
 & Uniform 0.75$\times$&N & 71 & 69.8 \\
 & APS & N & 64 & 69.0 \\
 & FSC & N & 62 & 70.2 \\
 & JCW & N & 59 & 72.2 \\
 & \textbf{PSS-Net} & \textbf{Y} & \textbf{58} & \textbf{72.3} \\
 & \textbf{PSS-Net} & \textbf{Y} & \textbf{46} & \textbf{70.9} \\
 & DMCP & N & 45 & 67.0 \\
 & JCW & N & 44 & 70.8 \\
 & DMCP & N & 43 & 66.1 \\
 & Uniform 0.5$\times$&N & 41 & 65.4 \\
 & JCW & N & 40 & 69.9 \\
 & \textbf{PSS-Net} & \textbf{Y} & \textbf{39} & \textbf{70.1} \\
\hline
\end{tabular}
}

}
\end{table}

{\color{black}
\textbf{Latency on Mobile Devices}.
Following JCW~\cite{zhao2022joint}, we also compare the TFLite latency on mobile devices equipped with ARM A72 single core CPU. Table\,\ref{cpu} provides the comparison results with JCW~\cite{zhao2022joint}, USNet~\cite{yu2019universally}, AutoSlim~\cite{yu2019autoslim}, AMC~\cite{he2018amc}, FSC~\cite{elsen2020fast}, DMCP~\cite{guo2020dmcp} and APS~\cite{wang2020revisiting}. The advantage of our PSS-Net is two-fold:
First, PSS-Net enjoys a weight-shared mechanism, leading to less storage requirement when deploying multiple subnets.
Second, under similar or even less latency, PSS-Net results in better performance in most cases.
Therefore, the applicability of PSS-Net on mobile devices is also well demonstrated.
}

{\color{black}
\textbf{Storage-constrained Devices.}
In addition to dynamically adapting to different FLOPs, CPU latency, and GPU latency constraints, as in the compact network methods, our PSS-Net can also be applied in storage resource-constrained scenarios by simply setting the type of resource constraints to the number of parameters at training time. Table\,\ref{tab:params_acc} shows the performance of different subnets with different number of parameters in the same PSS-Net supernet, which can be deployed independently on devices with different requirements of storage constraints.
}

\textbf{Complexity Comparison}.
We further compare the storage and training costs of PSS-Net, resource-adaptive methods~\cite{yu2019universally,yang2020mutualnet}, and the compact methods of model pruning. Denote the training cost of the supernet as $O_T$ and the storage cost as $O_S$. Given $N$ subnets, we summarize the cost complexity in Table\,\ref{tab:cost}.
For resource-adaptive networks, $k$ subnets of different sizes ($k=4$ for USNet and MutualNet, $k=3$ for PSS-Net) are trained in each batch, and thus the total training cost is not more than $kO_T$.
The resource-adaptive network methods require to store the parameters of means and variances of the BN layers for each of the $N$ subnets (the number of BN parameters for each MobileNet-V1/V2 subnet is about 1\% of the supernet's parameters $O_S$), and thus the total storage overhead is about $(1+0.01N)O_S$.
Most of the model pruning methods require a complete training for each compact subnet in isolation, so the total training cost is about $NO_T$ and the total storage cost is $NO_S$.
Taking $N = 100$ as an example, our PSS-Net achieves better performance than the model pruning methods (Table\,\ref{tab:pruning}) with about $3\%$ of the training cost and about $2\%$ of the storage cost compared to the model pruning methods (Table\,\ref{tab:cost}).
Note that the MDP\cite{liu2020joint} prunes model in three dimensions: channel, resolution and depth. In contrast, the subnet search space of PSS-Net only has two dimensions: resolution and channels. Nevertheless, the performance of the PSS-Net still exceeds the compact model trained by MDP, further demonstrating the effectiveness of PSS-Net.

\begin{table}[!t]
\centering
{\color{black}
\caption{Performance of PSS-Net subnets with different parameters. The backbone is MobileNet-V1.}
\vspace{-1.0em}
\label{tab:params_acc}
\begin{tabular}{c|cc} 
\hline
Model & Parameters (K) & Acc@1 (\%) \\ 
\hline
Subnet$_\text{A}$ & 4198 & 74.2 \\
Subnet$_\text{B}$ & 3667 & 73.8 \\
Subnet$_\text{C}$ & 3560 & 73.6 \\
Subnet$_\text{D}$ & 3006 & 68.1 \\
Subnet$_\text{E}$ & 2585 & 67.9 \\
\hline
\end{tabular}
}
\end{table}

\begin{table}[!t]
\centering
\caption{Complexity comparison of training and storage costs among existing resource-adaptive methods~\cite{yu2019universally,yang2020mutualnet}, model pruning, and our PSS-Net.}
\vspace{-1.0em}
\label{tab:cost}
\setlength{\tabcolsep}{1.9mm}
{
\begin{tabular}{cccc} 
\hline
Method & Subnet Num. &Training &Storage \\ 
\hline
USNet & $N$ & $<4O_T$ & $(1+0.01N)O_S$ \\
MutualNet & $N$ & $<4O_T$ & $(1+0.01N)O_S$ \\
Pruning & $N$ & $NO_T$ & $NO_S$ \\
PSS-Net & $N$ & $<3O_T$ & $(1+0.01N)O_S$ \\
\hline
\end{tabular}
}
\end{table}

\begin{table}[!t]
\centering
\caption{Comparison with a random search.}
\vspace{-1.0em}
\label{tab:compare_random}
\begin{tabular}{c|cc|c} 
\hline
\diagbox{FLOPs}{Acc@1(\%)}{Method} & Random & PSS-Net & Boost \\ 
\hline
300M & 71.2 & 72.3 & +1.1 \\
400M & 72.5 & 72.9 & +0.4 \\
500M & 73.7 & 73.8 & +0.1 \\
\hline
\end{tabular}
\end{table}

\subsection{Ablation Studies}\label{sec:ablation}
We further conduct ablation studies to analyze the effectiveness of our PSS-Net. All the experiments are conducted using MobileNet-V1 as the supernet which is trained under the FLOPs constraint.

\textbf{Compare with a Random Search.}
We compare PSS-Net with a random search under the same subnet space with resource constraints of 300M, 400M and 500M. As shown in Table\,\ref{tab:compare_random}, PSS-Net improves over a random search for different sizes of resource constraints, especially for smaller FLOPs constraints. This is due to the fact that the selection space of subnets with small FLOPs tends to be larger than that of large FLOPs, so that random searches are less likely to find better subnet structures.
It also illustrates the effectiveness of PSS-Net in using subnet pools to collect high-quality subnets during training.

\textbf{Low Bound of Subnet Space.}
Recall that we set $o_i^{max}$ to be the same as the channel number of the $i$-th layer in the backbone network, and the low bound of subnets $o_i^{min} = 0.75 \cdot o_i^{max}$. 
In this subsection, we consider different low bounds adopted by MutualNet~\cite{yang2020mutualnet} and show the experimental results in Table\,\ref{tab:compare_mutualnet}. Our PSS-Net shows better performance when the low bounds are set to different scales.

\begin{table}[!t]
\centering
\caption{Performance comparison under different low bounds of subnets.}
\vspace{-1.0em}
\label{tab:compare_mutualnet}
\setlength{\tabcolsep}{1.9mm}
\begin{tabular}{ccccc} 
\hline
Backbone & \begin{tabular}[c]{@{}c@{}}Lower\\Bound\end{tabular} & Method & \begin{tabular}[c]{@{}c@{}}FLOPs\\(M)\end{tabular} & \begin{tabular}[c]{@{}c@{}}Acc@1\\(\%)\end{tabular} \\ 
\hline
\multirow{4}{*}{MobileNet-V1} & \multirow{4}{*}{0.25} & \multirow{2}{*}{MutualNet} & 13 & 45.5 \\
 &  &  & 569 & 72.4 \\ 
\cline{3-5}
 &  & \multirow{2}{*}{PSS-Net} & 13 & \textbf{50.8 (+5.3)} \\
 &  &  & 569 & \textbf{72.8 (+0.4)} \\ 
\hline
\multirow{8}{*}{MobileNet-V2} & \multirow{4}{*}{0.7} & \multirow{2}{*}{MutualNet} & 57 & 64.2 \\
 &  &  & 301 & 72.9 \\ 
\cline{3-5}
 &  & \multirow{2}{*}{PSS-Net} & 57 & \textbf{65.4 (+1.2)} \\
 &  &  & 301 & \textbf{73.1 (+0.2)} \\ 
\cline{2-5}
 & \multirow{4}{*}{0.8} & \multirow{2}{*}{MutualNet} & 73 & 66.0 \\
 &  &  & 301 & 73.2 \\ 
\cline{3-5}
 &  & \multirow{2}{*}{PSS-Net} & 73 & \textbf{66.8 (+0.8)} \\
 &  &  & 301 & \textbf{73.5 (+0.3)} \\
\hline
\end{tabular}
\end{table}

\textbf{BN Calibration.}
Our PSS-Net follows the slimmable training paradigm which requires BN calibration for each subnet. However, the large volume of subnet space disables one-by-one calibration. Luckily, our introduced subnet pool encourages to collect more high-quality subnets. Consequently, we only need to calibrate $k$ subnets within top-$k$ performance metrics, and then pick the one with the best accuracy.
To verify this, in Table\,\ref{tab:topk}, we choose different values of $k$ and show the best subnets after BN calibration, which are constrained by the FLOPs of 300M, 400M and 500M. When $k \ge 2$, better subnets can be observed than simply considering the top-1. However, the performance gains are limited when $k > 5$, which indicates the best subnets fall into these subnets within top-5 performance metrics. Thus, we pick up 5 subnets in each pool for BN calibration, which is also our setting in all the experiments.

\begin{table}[!t]
\centering
\caption{BN calibration for subnets within top-$k$ performance metrics.}
\vspace{-1.0em}
\label{tab:topk}
\setlength{\tabcolsep}{1.3mm}
{
\begin{tabular}{c|c|ccc} 
\hline
\multirow{2}{*}{Pool} & \multicolumn{4}{c}{Acc@1(\%)} \\ 
\cline{2-5}
 & k=1 & k=5 & k=10 & k=20 \\ 
\hline
300M & 71.4 & 72.3 (+0.9) & 72.3 (+0.9) & 72.4 (+1.0) \\
400M & 72.7 & 72.9 (+0.2) & 72.9 (+0.2) & 72.9 (+0.2) \\
500M & 73.7 & 73.8 (+0.1) & 73.8 (+0.1) & 73.8 (+0.1) \\ 
\hline
Avg. Boost & / & +0.4 & +0.4 & +0.43 \\
\hline
\end{tabular}
}
\end{table}

\textbf{Other Hyper-parameters.}
We also perform an ablation study for the hyper-parameters $p_{end}$, $\eta_{end}$, $M$ {\color{black}and $divisor$} in the algorithms of \textbf{PSS-Net Training} and \textbf{Prioritized Subnet Sampling}. As shown in Table\,\ref{tab:hyper}, $p_{end}$ is insensitive to the hyper-parameter value and $\eta_{end}$ is slightly better given 0.01. For simplicity, we use $p_{end}=\eta_{end}=0.01$ in our experiments. 
It can also be seen from Table\,\ref{tab:hyper} that the performance of the subnet increases with the subnet pool size $M$, but using too large a subnet pool size leads to too much storage overhead, so we take $M=50$ in this paper.
{\color{black}
As for the channel divisor, a smaller divisor results in better performance. This is due to the fact that a smaller divisor leads to a larger subnet sampling space, which improves the subnet performance. As mentioned in Sec.\,\ref{sec:preliminary}, we choose $divisor=8$ to align channel number with the warp size of most mobile CPUs for more efficient computation.
}

\textbf{Subnet Fine-tuning.}
As mentioned in the introduction section, existing researches suffer the problem of subnet mismatch, which causes insufficient training of high-quality subnets. Our PSS-Net solves this problem by giving priority to sampling high-quality subnets in the subnet pool. In this subsection, we further investigate whether the best subnets have been well trained in PSS-Net. To this end, we select the best subnets from the 300M-, 400M-, and 500M-subnet pools, and further fine-tune these subnets for 25 epochs. Table\,\ref{tab:fine-tune} shows the performance comparison before/after fine-tuning. As can be seen, fine-tuning incurs slight performance drops in these three cases. This experiment demonstrates that our high-quality subnets are already well trained and our PSS-Net well solves the subnet mismatch problems.

\begin{table}[!t]
\centering
\caption{
Ablation studies of the hyper-parameters $p_{end}$, $\eta_{end}$, $M$, {\color{black} and $divisor$} in the algorithms of \textbf{PSS-Net Training} and \textbf{Prioritized Subnet Sampling}.
}
\vspace{-1.0em}
\label{tab:hyper}
\setlength{\tabcolsep}{0.55mm}{
\begin{tabular}{c|c|cc|cc|cc|cc} 
\hline
\multirow{3}{*}{Pool} & \multicolumn{9}{c}{Acc@1(\%)} \\ 
\cline{2-10}
 & baseline & \multicolumn{2}{c|}{$p_{end}$} & \multicolumn{2}{c|}{$\eta_{end}$} & \multicolumn{2}{c|}{$M$} & \multicolumn{2}{c}{$divisor$} \\ 
\cline{2-10}
 & \begin{tabular}[c]{@{}c@{}}$p_{end}=0.01$\\$\eta_{end}=0.01$\\$M=50$\\$divisor=8$\end{tabular} & 0.1 & 0.001 & 0.1 & 0.001 & 10 & 500 & {4} & {16} \\ 
\hline
300M & 72.3 & 72.2 & 72.2 & 72.1 & 72.2 & 72.0 & 72.5 & {72.5} & {71.2} \\
400M & 72.9 & 72.7 & 73.0 & 72.6 & 72.9 & 72.7 & 73.0 & {72.9} & {72.6} \\
500M & 73.8 & 73.9 & 73.7 & 73.7 & 73.6 & 73.5 & 73.8 & {73.9} & {73.8} \\ 
\hline
Avg. Boost & / & -0.1 & +0.0 & -0.2 & -0.1 & -0.3 & +0.1 & {+0.1} & {-0.47} \\
\hline
\end{tabular}
}
\end{table}

\begin{table}[!t]
\centering
\caption{Performance comparison before/after fine-tuning the best subnets.}
\vspace{-1.0em}
\label{tab:fine-tune}
\begin{tabular}{cccc} 
\hline
Subnet & FLOPs & Acc$_\textit{PSS}$ (\%) & Acc$_\textit{Fine-tune}$ (\%) \\ 
\hline
A & 300M & \textbf{72.3} & 71.9 (-0.4) \\
B & 400M & \textbf{72.9} & 72.7 (-0.2) \\
C & 500M & \textbf{73.8} & 73.5 (-0.3) \\
\hline
\end{tabular}
\end{table}

\begin{table}[!t]
\centering
{\color{black}
\caption{Performance of Faster-RCNN on COCO with ResNet-50 backbone.}
\vspace{-1.0em}
\label{coco}
}
{\color{black}

\begin{tabular}{cccc} 
\hline
Method & FLOPs (G) & Box~mAP (\%) & $\Delta$Box~mAP (\%) \\ 
\hline
Uniform 1.0$\times$ & 206 & 36.4 & +0.0 \\
USNet & 206 & 36.8 & +0.4 \\
MutualNet & 186 & 36.8 & +0.4 \\
\textbf{PSS-Net} & \textbf{174~} & \textbf{36.8} & \textbf{+0.4} \\ 
\hline
Uniform 0.75$\times$ & 170 & 34.7 & +0.0 \\
USNet & 170 & 36.3 & +1.6 \\
MutualNet & 167 & 36.6 & +1.9 \\
\textbf{PSS-Net} & \textbf{165} & \textbf{36.7} & \textbf{+2.0} \\ 
\hline
Uniform 0.5$\times$ & 144 & 32.7 & +0.0 \\
USNet & 144 & 34.5 & +1.8 \\
MutualNet & 143 & 34.8 & +2.4 \\
\textbf{PSS-Net} & \textbf{142} & \textbf{36.0} & \textbf{+3.3} \\ 
\hline
Uniform 0.25$\times$ & 129 & 27.5 & +0.0 \\
USNet & 129 & 30.4 & +2.9 \\
MutualNet & 126 & 31.8 & +4.3 \\
\textbf{PSS-Net} & \textbf{126} & \textbf{35.6} & \textbf{+8.1} \\
\hline
\end{tabular}
}
\end{table}

{\color{black}
\subsection{Detection Performance}
We further verify the detection performance on COCO~\cite{lin2014microsoft} using Faster-RCNN~\cite{girshick2015fast}. The network backbone is ResNet-50. Table\,\ref{coco} manifests the experiment comparison. We observe similar conclusion as in classification where the proposed PSS-Net results in subnets that have similar or less computation cost, but significant increase in mAP performance when comparing to existing USNet and MutualNet. The superiority is outstanding in particular to a smaller backbone such as $0.25\times$. Therefore, the efficacy of our PSS-Net is well demonstrated even upon different tasks.
}

\section{Limitations}
We construct lookup tables based on the block latency of a supernet for a quick calculation of resource consumption in PSS-Net training. Though effective enough, the predicted latency sometimes still deviates slightly from the real latency, which would misguide the sample of subnets. Besides, theoretical analyses on why the proposed subnet sampling works well and the relationship between the subnet pools and subnet design space remain to be further explored.

\section{Conclusion}\label{conclusion}
We have proposed Prioritized Subnet Sampling (PSS-Net) to train a resource-adaptive supernet. Compared to existing resource-adaptive networks, our PSS-Net offers subnets with a better accuracy-resource trade-off by maintaining subnet pools, each of which is conditioned on a particular resource constraint, and assigning priority to high-quality subnets for training. Our PSS-Net produces a set of high-quality subnets with different resource consumptions, thus allowing a fast switch of high-quality subnets for inference when the available resources vary.
Besides, compared to the compact network methods such as pruning and NAS, our PSS-Net significantly reduces the training and storage cost.

\section*{Acknowledgments}
This work was supported by the National Science Fund for Distinguished Young Scholars (No. 62025603), the National Natural Science Foundation of China (No. U21B2037, No. U22B2051, No. 62176222, No. 62176223, No. 62176226, No. 62072386, No. 62072387, No. 62072389, No. 62002305 and No. 62272401), Guangdong Basic and Applied Basic Research Foundation (No. 2019B1515120049), and the Natural Science Foundation of Fujian Province of China (No. 2021J01002,  No. 2022J06001).

\appendices




\ifCLASSOPTIONcaptionsoff
  \newpage
\fi




\bibliographystyle{IEEEtran}
\bibliography{main}

\begin{thebibliography}{10}
\providecommand{\url}[1]{#1}
\csname url@samestyle\endcsname
\providecommand{\newblock}{\relax}
\providecommand{\bibinfo}[2]{#2}
\providecommand{\BIBentrySTDinterwordspacing}{\spaceskip=0pt\relax}
\providecommand{\BIBentryALTinterwordstretchfactor}{4}
\providecommand{\BIBentryALTinterwordspacing}{\spaceskip=\fontdimen2\font plus
\BIBentryALTinterwordstretchfactor\fontdimen3\font minus
  \fontdimen4\font\relax}
\providecommand{\BIBforeignlanguage}[2]{{%
\expandafter\ifx\csname l@#1\endcsname\relax
\typeout{** WARNING: IEEEtran.bst: No hyphenation pattern has been}%
\typeout{** loaded for the language `#1'. Using the pattern for}%
\typeout{** the default language instead.}%
\else
\language=\csname l@#1\endcsname
\fi
#2}}
\providecommand{\BIBdecl}{\relax}
\BIBdecl

\bibitem{ruan2021dpfps}
X.~Ruan, Y.~Liu, B.~Li, C.~Yuan, and W.~Hu, ``Dpfps: Dynamic and progressive
  filter pruning for compressing convolutional neural networks from scratch,''
  in \emph{Proceedings of the AAAI Conference on Artificial Intelligence
  (AAAI)}, 2021, pp. 2495--2503.

\bibitem{wang2020pruning}
Y.~Wang, X.~Zhang, L.~Xie, J.~Zhou, H.~Su, B.~Zhang, and X.~Hu, ``Pruning from
  scratch,'' in \emph{Proceedings of the AAAI Conference on Artificial
  Intelligence (AAAI)}, 2020, pp. 12\,273--12\,280.

\bibitem{cai2019once}
H.~Cai, C.~Gan, T.~Wang, Z.~Zhang, and S.~Han, ``Once-for-all: Train one
  network and specialize it for efficient deployment,'' in \emph{Proceedings of
  the International Conference on Learning Representations (ICLR)}, 2020.

\bibitem{dai2019chamnet}
X.~Dai, P.~Zhang, B.~Wu, H.~Yin, F.~Sun, Y.~Wang, M.~Dukhan, Y.~Hu, Y.~Wu,
  Y.~Jia \emph{et~al.}, ``Chamnet: Towards efficient network design through
  platform-aware model adaptation,'' in \emph{Proceedings of the IEEE/CVF
  Conference on Computer Vision and Pattern Recognition (CVPR)}, 2019, pp.
  11\,398--11\,407.

\bibitem{yu2018slimmable}
J.~Yu, L.~Yang, N.~Xu, J.~Yang, and T.~Huang, ``Slimmable neural networks,'' in
  \emph{Proceedings of the International Conference on Learning Representations
  (ICLR)}, 2019.

\bibitem{liu2019metapruning}
Z.~Liu, H.~Mu, X.~Zhang, Z.~Guo, X.~Yang, K.-T. Cheng, and J.~Sun,
  ``Metapruning: Meta learning for automatic neural network channel pruning,''
  in \emph{Proceedings of the IEEE/CVF International Conference on Computer
  Vision (ICCV)}, 2019, pp. 3296--3305.

\bibitem{yang2020mutualnet}
T.~Yang, S.~Zhu, C.~Chen, S.~Yan, M.~Zhang, and A.~Willis, ``Mutualnet:
  Adaptive convnet via mutual learning from network width and resolution,'' in
  \emph{Proceedings of the European Conference on Computer Vision (ECCV)},
  2020, pp. 299--315.

\bibitem{yu2019universally}
J.~Yu and T.~S. Huang, ``Universally slimmable networks and improved training
  techniques,'' in \emph{Proceedings of the IEEE/CVF International Conference
  on Computer Vision (ICCV)}, 2019, pp. 1803--1811.

\bibitem{lou2021dynamic}
W.~Lou, L.~Xun, A.~Sabet, J.~Bi, J.~Hare, and G.~V. Merrett, ``Dynamic-ofa:
  Runtime dnn architecture switching for performance scaling on heterogeneous
  embedded platforms,'' in \emph{Proceedings of the IEEE/CVF Conference on
  Computer Vision and Pattern Recognition (CVPR)}, 2021, pp. 3110--3118.

\bibitem{yu2020bignas}
J.~Yu, P.~Jin, H.~Liu, G.~Bender, P.-J. Kindermans, M.~Tan, T.~Huang, X.~Song,
  R.~Pang, and Q.~Le, ``Bignas: Scaling up neural architecture search with big
  single-stage models,'' in \emph{Proceedings of the European Conference on
  Computer Vision (ECCV)}, 2020, pp. 702--717.

\bibitem{wang2021attentivenas}
D.~Wang, M.~Li, C.~Gong, and V.~Chandra, ``Attentivenas: Improving neural
  architecture search via attentive sampling,'' in \emph{Proceedings of the
  IEEE/CVF Conference on Computer Vision and Pattern Recognition (CVPR)}, 2021,
  pp. 6418--6427.

\bibitem{yu2019autoslim}
J.~Yu and T.~Huang, ``Autoslim: Towards one-shot architecture search for
  channel numbers,'' \emph{arXiv preprint arXiv:1903.11728}, 2019.

\bibitem{lin2020channel}
M.~Lin, R.~Ji, Y.~Zhang, B.~Zhang, Y.~Wu, and Y.~Tian, ``Channel pruning via
  automatic structure search,'' in \emph{Proceedings of the International Joint
  Conference on Artificial Intelligence (IJCAI)}, 2020, pp. 673 -- 679.

\bibitem{guo2020single}
Z.~Guo, X.~Zhang, H.~Mu, W.~Heng, Z.~Liu, Y.~Wei, and J.~Sun, ``Single path
  one-shot neural architecture search with uniform sampling,'' in
  \emph{Proceedings of the European Conference on Computer Vision (ECCV)},
  2020, pp. 544--560.

\bibitem{sahni2021compofa}
M.~Sahni, S.~Varshini, A.~Khare, and A.~Tumanov, ``Compofa: Compound
  once-for-all networks for faster multi-platform deployment,'' in
  \emph{Proceedings of the International Conference on Learning Representations
  (ICLR)}, 2021.

\bibitem{howard2017mobilenets}
A.~G. Howard, M.~Zhu, B.~Chen, D.~Kalenichenko, W.~Wang, T.~Weyand,
  M.~Andreetto, and H.~Adam, ``Mobilenets: Efficient convolutional neural
  networks for mobile vision applications,'' \emph{arXiv preprint
  arXiv:1704.04861}, 2017.

\bibitem{sandler2018mobilenetv2}
M.~Sandler, A.~Howard, M.~Zhu, A.~Zhmoginov, and L.-C. Chen, ``Mobilenetv2:
  Inverted residuals and linear bottlenecks,'' in \emph{Proceedings of the
  IEEE/CVF Conference on Computer Vision and Pattern Recognition (CVPR)}, 2018,
  pp. 4510--4520.

\bibitem{he2016deep}
K.~He, X.~Zhang, S.~Ren, and J.~Sun, ``Deep residual learning for image
  recognition,'' in \emph{Proceedings of the IEEE/CVF Conference on Computer
  Vision and Pattern Recognition (CVPR)}, 2016, pp. 770--778.

\bibitem{guo2020channel}
J.~Guo, W.~Ouyang, and D.~Xu, ``Channel pruning guided by classification loss
  and feature importance,'' in \emph{Proceedings of the AAAI Conference on
  Artificial Intelligence (AAAI)}, 2020, pp. 10\,885--10\,892.

\bibitem{li2020eagleeye}
B.~Li, B.~Wu, J.~Su, and G.~Wang, ``Eagleeye: Fast sub-net evaluation for
  efficient neural network pruning,'' in \emph{Proceedings of the European
  Conference on Computer Vision (ECCV)}, 2020, pp. 639--654.

\bibitem{he2018amc}
Y.~He, J.~Lin, Z.~Liu, H.~Wang, L.-J. Li, and S.~Han, ``Amc: Automl for model
  compression and acceleration on mobile devices,'' in \emph{Proceedings of the
  European conference on computer vision (ECCV)}, 2018, pp. 784--800.

\bibitem{han2015learning}
S.~Han, J.~Pool, J.~Tran, and W.~Dally, ``Learning both weights and connections
  for efficient neural network,'' in \emph{Proceedings of the Advances in
  Neural Information Processing Systems (NeurIPS)}, 2015, pp. 1135--1143.

\bibitem{lin2020hrank}
M.~Lin, R.~Ji, Y.~Wang, Y.~Zhang, B.~Zhang, Y.~Tian, and L.~Shao, ``Hrank:
  Filter pruning using high-rank feature map,'' in \emph{Proceedings of the
  IEEE/CVF Conference on Computer Vision and Pattern Recognition (CVPR)}, 2020,
  pp. 1529--1538.

\bibitem{frankle2018lottery}
J.~Frankle and M.~Carbin, ``The lottery ticket hypothesis: Finding sparse,
  trainable neural networks,'' in \emph{Proceedings of the International
  Conference on Learning Representations (ICLR)}, 2018.

\bibitem{zhao2022joint}
T.~Zhao, X.~S. Zhang, W.~Zhu, J.~Wang, S.~Yang, J.~Liu, and J.~Cheng, ``Joint
  channel and weight pruning for model acceleration on moblie devices,'' in
  \emph{Proceedings of the European Conference on Computer Vision (ECCV)},
  2022.

\bibitem{guo2020multi}
J.~Guo, W.~Ouyang, and D.~Xu, ``Multi-dimensional pruning: A unified framework
  for model compression,'' in \emph{Proceedings of the IEEE/CVF Conference on
  Computer Vision and Pattern Recognition (CVPR)}, 2020, pp. 1508--1517.

\bibitem{wang2022versatile}
H.~Wang, Y.~Zhang, and J.~Wu, ``Versatile, full-spectrum, and swift network
  sampling for model generation,'' \emph{Pattern Recognition}, vol. 129, p.
  108729, 2022.

\bibitem{lin2017runtime}
J.~Lin, Y.~Rao, J.~Lu, and J.~Zhou, ``Runtime neural pruning,'' in
  \emph{Proceedings of the Advances in Neural Information Processing Systems
  (NeurIPS)}, 2017, pp. 2178--2188.

\bibitem{bejnordi2019batch}
B.~E. Bejnordi, T.~Blankevoort, and M.~Welling, ``Batch-shaping for learning
  conditional channel gated networks,'' in \emph{Proceedings of the
  International Conference on Learning Representations (ICLR)}, 2020.

\bibitem{huang2017multi}
G.~Huang, D.~Chen, T.~Li, F.~Wu, L.~van~der Maaten, and K.~Q. Weinberger,
  ``Multi-scale dense networks for resource efficient image classification,''
  in \emph{Proceedings of the International Conference on Learning
  Representations (ICLR)}, 2018.

\bibitem{wang2018skipnet}
X.~Wang, F.~Yu, Z.-Y. Dou, T.~Darrell, and J.~E. Gonzalez, ``Skipnet: Learning
  dynamic routing in convolutional networks,'' in \emph{Proceedings of the
  European Conference on Computer Vision (ECCV)}, 2018, pp. 409--424.

\bibitem{berman2020aows}
M.~Berman, L.~Pishchulin, N.~Xu, M.~B. Blaschko, and G.~Medioni, ``Aows:
  Adaptive and optimal network width search with latency constraints,'' in
  \emph{Proceedings of the IEEE/CVF Conference on Computer Vision and Pattern
  Recognition (CVPR)}, 2020, pp. 11\,217--11\,226.

\bibitem{deng2009imagenet}
J.~Deng, W.~Dong, R.~Socher, L.-J. Li, K.~Li, and L.~Fei-Fei, ``Imagenet: A
  large-scale hierarchical image database,'' in \emph{Proceedings of the
  IEEE/CVF Conference on Computer Vision and Pattern Recognition (CVPR)}, 2009,
  pp. 248--255.

\bibitem{liu2020joint}
Z.~Liu, X.~Zhang, Z.~Shen, Z.~Li, Y.~Wei, K.-T. Cheng, and J.~Sun, ``Joint
  multi-dimension pruning,'' \emph{arXiv preprint arXiv:2005.08931}, 2020.

\bibitem{yang2018netadapt}
T.-J. Yang, A.~Howard, B.~Chen, X.~Zhang, A.~Go, M.~Sandler, V.~Sze, and
  H.~Adam, ``Netadapt: Platform-aware neural network adaptation for mobile
  applications,'' in \emph{Proceedings of the European Conference on Computer
  Vision (ECCV)}, 2018, pp. 285--300.

\bibitem{li2021towards}
Y.~Li, S.~Lin, J.~Liu, Q.~Ye, M.~Wang, F.~Chao, F.~Yang, J.~Ma, Q.~Tian, and
  R.~Ji, ``Towards compact cnns via collaborative compression,'' in
  \emph{Proceedings of the IEEE/CVF Conference on Computer Vision and Pattern
  Recognition (CVPR)}, 2021, pp. 6438--6447.

\bibitem{guo2020dmcp}
S.~Guo, Y.~Wang, Q.~Li, and J.~Yan, ``Dmcp: Differentiable markov channel
  pruning for neural networks,'' in \emph{Proceedings of the IEEE/CVF
  Conference on Computer Vision and Pattern Recognition (CVPR)}, 2020, pp.
  1539--1547.

\bibitem{luo2017thinet}
J.-H. Luo, J.~Wu, and W.~Lin, ``Thinet: A filter level pruning method for deep
  neural network compression,'' in \emph{Proceedings of the IEEE/CVF
  International Conference on Computer Vision (ICCV)}, 2017, pp. 5058--5066.

\bibitem{he2019filter}
Y.~He, P.~Liu, Z.~Wang, Z.~Hu, and Y.~Yang, ``Filter pruning via geometric
  median for deep convolutional neural networks acceleration,'' in
  \emph{Proceedings of the IEEE/CVF Conference on Computer Vision and Pattern
  Recognition (CVPR)}, 2019, pp. 4340--4349.

\bibitem{luo2020neural}
J.-H. Luo and J.~Wu, ``Neural network pruning with residual-connections and
  limited-data,'' in \emph{Proceedings of the IEEE/CVF Conference on Computer
  Vision and Pattern Recognition (CVPR)}, 2020, pp. 1458--1467.

\bibitem{gao2020discrete}
S.~Gao, F.~Huang, J.~Pei, and H.~Huang, ``Discrete model compression with
  resource constraint for deep neural networks,'' in \emph{Proceedings of the
  IEEE/CVF Conference on Computer Vision and Pattern Recognition (CVPR)}, 2020,
  pp. 1899--1908.

\bibitem{liu2019darts}
H.~Liu, K.~Simonyan, and Y.~Yang, ``Darts: Differentiable architecture
  search,'' in \emph{Proceedings of the International Conference on Learning
  Representations (ICLR)}, 2019.

\bibitem{liu2018progressive}
C.~Liu, B.~Zoph, M.~Neumann, J.~Shlens, W.~Hua, L.-J. Li, L.~Fei-Fei,
  A.~Yuille, J.~Huang, and K.~Murphy, ``Progressive neural architecture
  search,'' in \emph{Proceedings of the European conference on computer vision
  (ECCV)}, 2018, pp. 19--34.

\bibitem{zoph2018learning}
B.~Zoph, V.~Vasudevan, J.~Shlens, and Q.~V. Le, ``Learning transferable
  architectures for scalable image recognition,'' in \emph{Proceedings of the
  IEEE/CVF Conference on Computer Vision and Pattern Recognition (CVPR)}, 2018,
  pp. 8697--8710.

\bibitem{elsen2020fast}
E.~Elsen, M.~Dukhan, T.~Gale, and K.~Simonyan, ``Fast sparse convnets,'' in
  \emph{Proceedings of the IEEE/CVF Conference on Computer Vision and Pattern
  Recognition (CVPR)}, 2020, pp. 14\,629--14\,638.

\bibitem{wang2020revisiting}
J.~Wang, H.~Bai, J.~Wu, X.~Shi, J.~Huang, I.~King, M.~Lyu, and J.~Cheng,
  ``Revisiting parameter sharing for automatic neural channel number search,''
  in \emph{Proceedings of the Advances in Neural Information Processing Systems
  (NeurIPS)}, vol.~33, 2020, pp. 5991--6002.

\bibitem{lin2014microsoft}
T.-Y. Lin, M.~Maire, S.~Belongie, J.~Hays, P.~Perona, D.~Ramanan,
  P.~Doll{\'a}r, and C.~L. Zitnick, ``Microsoft coco: Common objects in
  context,'' in \emph{Proceedings of the European Conference on Computer Vision
  (ECCV)}, 2014, pp. 740--755.

\bibitem{girshick2015fast}
R.~Girshick, ``Fast r-cnn,'' in \emph{Proceedings of the IEEE/CVF Conference on
  Computer Vision and Pattern Recognition (CVPR)}, 2015, pp. 1440--1448.

\end{thebibliography}

%



%

\begin{IEEEbiography}[{\includegraphics[width=1in,height=1.25in,clip,keepaspectratio]{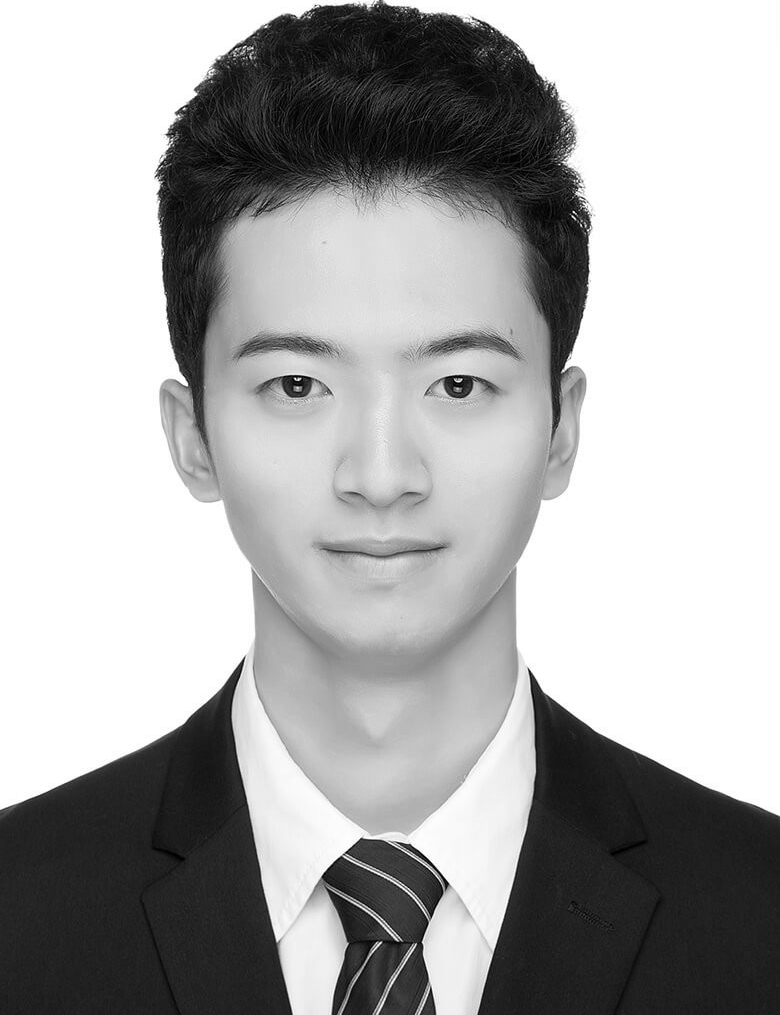}}]{Bohong Chen} received the B.E. degree in Computer Science, School of Informatics, Xiamen University, Xiamen, China, in 2020.
He is currently working toward the master’s degree from Xiamen University, China.
His research interests include computer vision, and neural network compression \& acceleration.
\end{IEEEbiography}

\begin{IEEEbiography}[{\includegraphics[width=1in,height=1.25in,clip,keepaspectratio]{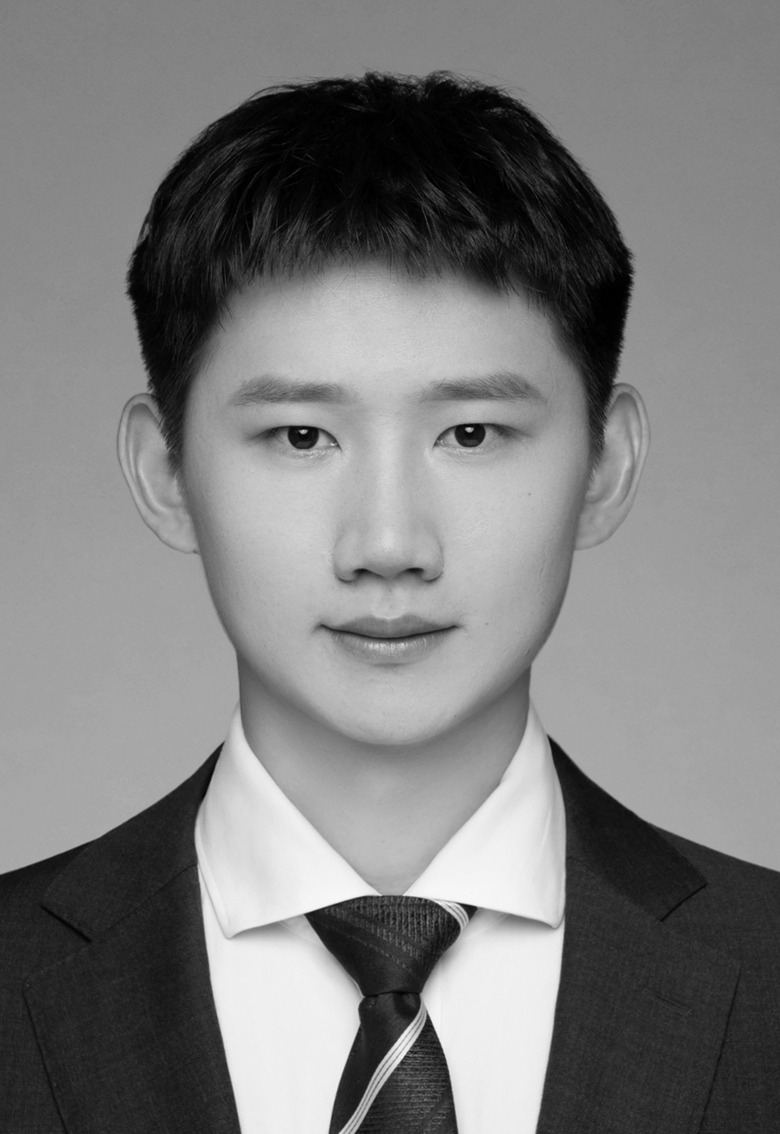}}]{Mingbao Lin} finished his M.S.-Ph.D. study and obtained the Ph.D. degree in intelligence science and technology from Xiamen University, Xiamen, China, in 2022. Earlier, he received the B.S. degree from Fuzhou University, Fuzhou, China, in 2016.

He is currently a senior researcher with the Tencent Youtu Lab, Shanghai, China. His publications on top-tier conferences/journals include IEEE TPAMI, IJCV, IEEE TIP, IEEE TNNLS, CVPR, NeurIPS, AAAI, IJCAI, ACM MM and so on. His current research interest is to develop efficient vision model, as well as information retrieval.
\end{IEEEbiography}

\begin{IEEEbiography}[{\includegraphics[width=1in,height=1.25in,clip,keepaspectratio]{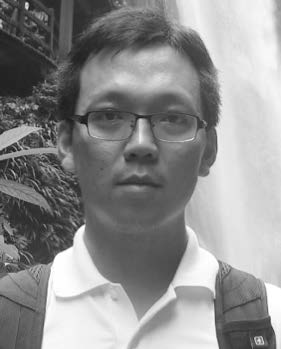}}]{Rongrong Ji}
(Senior Member, IEEE) is currently a Professor and the Director of the Intelligent Multimedia Technology Laboratory, and the Dean Assistant with the School of Information Science and Engineering, Xiamen University, Xiamen, China. His work mainly focuses on innovative technologies for multimedia signal processing, computer vision, and pattern recognition, with over 100 papers published in international journals and conferences. He is a member of the ACM. He was a recipient of the ACM Multimedia Best Paper Award and the Best Thesis Award of Harbin Institute of Technology. He serves as an Associate/Guest Editor for international journals and magazines such as \emph{Neurocomputing}, \emph{Signal Processing}, \emph{Multimedia Tools and Applications}, the \emph{IEEE Multimedia Magazine}, and the \emph{Multimedia Systems}. He also serves as program committee member for several Tier-$1$ international conferences.
\end{IEEEbiography}

\begin{IEEEbiography}[{\includegraphics[width=1in,height=1.25in,clip,keepaspectratio]{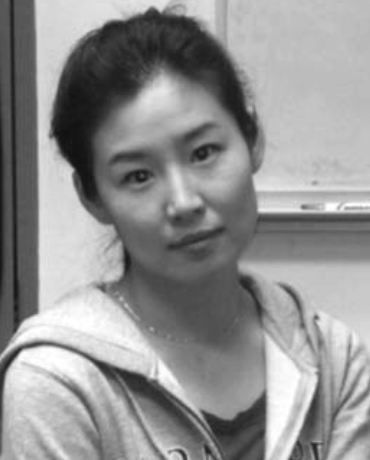}}]{Liujuan Cao}
received the B.S., M.S., and Ph.D degrees from the School of Computer Science and Technology, Harbin Engineering University. She is currently and associate professor at Xiamen University. Her research interests are computer vision and pattern recognition. She has authored over 40 papers in top and major tired journals and conferences, including CVPR, TIP, etc. She is the Financial Chair of the IEEE MMSP 2015, the Workshop Chair of the ACM ICIMCS 2016, and the Local Chair of the Visual and Learning Seminar 2017.
\end{IEEEbiography}







\end{document}